\documentclass[preprint,12pt,authoryear]{elsarticle}
\usepackage[margin=3.0cm]{geometry}

\usepackage{amssymb}
\usepackage{amsmath}
\usepackage{subcaption}
\usepackage{booktabs}
\usepackage{siunitx}
\usepackage{hyperref}
\usepackage{array}
\usepackage{multirow}
\usepackage{lineno}

\usepackage{xcolor}
\usepackage{array}
\definecolor{rulegrey}{gray}{0.75}
\newcolumntype{?}{!{\color{rulegrey}\vrule width 0.4pt}}

\usepackage{colortbl}
\newcommand{\grouprule}{%
  \noalign{\vskip1.5pt}\arrayrulecolor{rulegrey}\cmidrule{1-8}\arrayrulecolor{black}\noalign{\vskip1.5pt}}

\hypersetup{
    colorlinks=true,
    linkcolor=blue,
    filecolor=magenta,      
    urlcolor=cyan,
    pdftitle={ALFRED: Requirement-driven development of an open-source mobile manipulator for long-term plant monitoring},
    pdfpagemode=FullScreen,
    }

\journal{}

\begin{document}

\begin{frontmatter}


\author[label1,label2]{Ciarán Miceal Johnson}
\ead{Ciaran.Johnson@ed.ac.uk}
\ead[url]{https://ciaranjohnson.github.io/}

\author[label2]{Christopher Quail}

\author[label1]{Garry Ellard}
\ead{Garry.Ellard@ed.ac.uk}
\ead[url]{https://uoe.sharepoint.com/sites/InformaticsMakerspace}

\author[label2]{Alistair McConnell\corref{cor1}}
\ead{Alistair.McConnell@hw.ac.uk}

\author[label1]{Steve Tonneau}
\ead{stonneau@exseed.ed.ac.uk}

\author[label3]{Fernando Auat Cheein\corref{cor1}}
\ead{FAuat@harper-adams.ac.uk}

\cortext[cor1]{Corresponding authors}

\affiliation[label1]{organization={The University of Edinburgh},
           addressline={Old College, South Bridge}, 
           city={Edinburgh},
           postcode={EH8 9YL}, 
           state={Scotland},
           country={UK}}

\affiliation[label2]{organization={Heriot-Watt University},
           addressline={Riccarton}, 
           city={Edinburgh},
           postcode={EH14 4AS}, 
           state={Scotland},
           country={UK}}

\affiliation[label3]{organization={Harper Adams University},
           addressline={Edgmond}, 
           city={Newport},
           postcode={TF10 8NB}, 
           state={England},
           country={UK}}

\title{ALFRED: Requirement-driven development of an open-source mobile manipulator for long-term plant monitoring} 

\begin{abstract}
Tracking seasonal change in crops and forests requires observing the same plants repeatedly. Ground robots can do this at close range, and a manipulator gives their sensors more viewpoints. Yet the robots behind long-term field datasets are rarely released with their design files, and how a robot's own structure limits arm reach and occludes its sensors is seldom compared between builds. We present ALFRED, an open-source mobile manipulator built from commercially available components. It carries a six-degree-of-freedom arm, LiDAR, RGB-D cameras, RTK GNSS and an IMU on an Ackermann-steered base, all mounted on a reconfigurable aluminium strut frame, and runs containerised ROS software. It was developed through four builds against six requirements for repeated outdoor deployment: durability, modularity, repairability, sensing reach, endurance and reproducibility. Model-based analysis of the last three builds shows the usable share of the arm's reachable poses rising from 34.0\% to 60.0\% and then 66.1\%, and ray casting shows that only the final build keeps the frame-mounted LiDAR's horizontal view clear both forwards and backwards. ALFRED completed a year of monthly forest surveys (528 traversals) without missing a scheduled collection. This was despite battery degradation, reconfiguration for another researcher's study, and the parallel development of ALFRED~2.0 for autonomous crop-row operation, with each switch between builds taking about six hours. The deployment also showed that mechanical modularity is only as dependable as the robot description that tracks it. Design files, bill of materials, robot descriptions, containers and build documentation are openly released.
\end{abstract}

\begin{graphicalabstract}
\includegraphics[width=\linewidth]{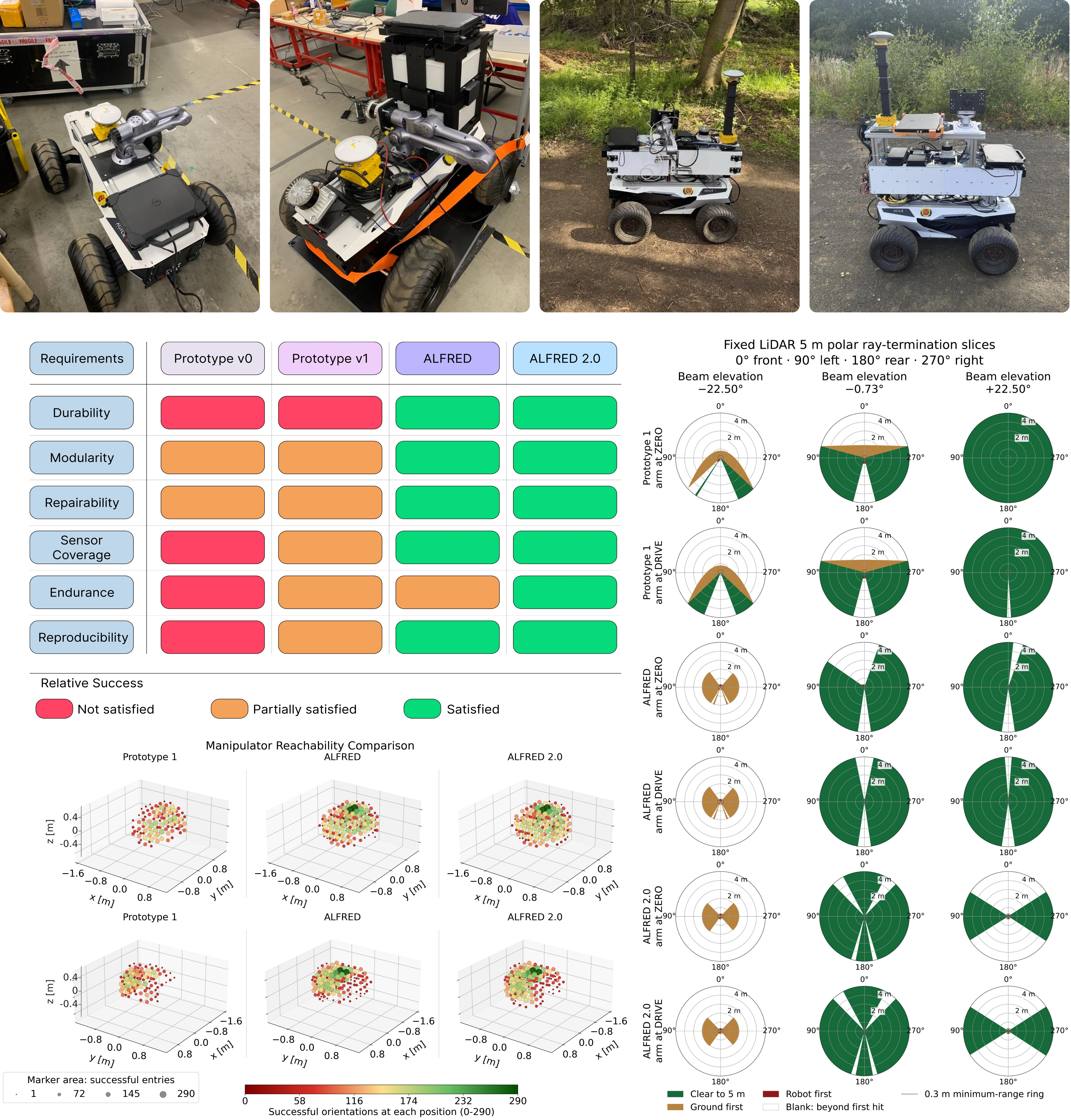}
\end{graphicalabstract}

\begin{highlights}
\item Four builds of a field mobile manipulator traced to six requirements
\item Usable arm workspace rose from 34\% to 66\% of reachable poses
\item Frame design and arm posture govern LiDAR self-occlusion
\item A year of monthly forest surveys with no missed collection
\item Design files, robot descriptions and software are open source
\end{highlights}


\end{frontmatter}


\section{Introduction}
\label{sec:alfred_intro}

Crop growth models simulate crop development from weather, soil and management practices, but numerous uncertainties compromise their performance \citep{dlamini2023assimReview}. Data assimilation can reduce this uncertainty using in-season measurements of crop traits \citep{huang2015laiWofost}. Acquiring those measurements conventionally requires manual and often destructive sampling, which is slow, labour-intensive and costly \citep{atefi2021RoboticPhenotyping}. Destructive sampling is also unsuitable for monitoring a crop continuously through its growth cycle: the plant is removed or damaged, and the gap it leaves alters the plants either side of it, which partially compensate for its loss \citep{pavek2006missingPlants}. Non-destructive measurement, by contrast, allows the same plants to be measured repeatedly and their growth tracked over time \citep{poorter2023pitfalls}.

Satellites and unmanned aerial vehicles (UAVs) both provide in-season observations for crop model data assimilation, and each makes a different trade-off between coverage and detail \citep{wang2024RSDAReview}. Satellites cover whole regions, but their resolution is coarse and cloud cover can leave long gaps between usable images during the growing season \citep{matese2015Intercomparison, whitcraft2015Cloud}.UAVs offer finer spatial resolution and more flexible revisits than satellites, at the cost of coverage \citep{bazrafkan2025UASSatellite}. However, they view the crop from above or at an oblique angle, so lower leaves, stems and internal canopy structure are often occluded in dense or tall crops \citep{zhang2026GroundRobotReview}. Low-altitude multi-rotor flight adds a further problem, since rotor downwash can deform flexible plants and alter the canopy geometry being measured \cite{shi2021UAVDownwash}. Unmanned ground vehicles (UGVs) do not suffer from the same issue and can sense plants at close range, where more information can be acquired about individual organs, and adding a manipulator allows cameras to be positioned and oriented at the best viewpoints \citep{atefi2021RoboticPhenotyping}. Because the end effector can be swapped, a single mobile manipulator can be reconfigured for different tasks \citep{guri2024Hefty}, allowing several kinds of data to be collected from the same plants. This article focuses on mobile manipulators of this kind.

Long-term experiments and data collection place additional demands on the platform. The platform must be able to operate outdoors all year round, be reconfigurable as sensing requirements change, and maintainable and repairable. Further demands arise when the platform is needed for other experiments, as happens when it is shared between projects, or during planned upgrades, which must be built and validated without missing scheduled data collections. Existing research shows that long-term deployment of agricultural robots is achievable \citep{bender2020Ladybird, polvara2024BLT}, as are modular platforms \citep{grimstad2017ThorvaldII, guri2024Hefty}. However, these studies rarely discuss the rapid prototyping required before and during sustained, repeated deployment, or compare successive prototype builds against desirable characteristics to quantify their improvements. Additionally, reachability analysis \citep{zacharias2007CapabilityMap, vahrenkamp2013ReachabilityInversion} and LiDAR placement \citep{kim2020LidarPlacement, hu2022MultiLidarPlacement} are common tools in manipulation and autonomous driving, but are rarely used as development tools to compare agricultural robot prototypes.

This paper presents the Autonomous Leaf and Forestry Research Exploration Device (ALFRED), a mobile manipulator assembled from commercially available components for repeated close-range sensing of crops and forests. It was developed through four builds, each motivated by a requirement its predecessor failed to meet, against six design requirements: outdoor durability, modularity and reconfigurability, repairability, sensing reach and coverage, endurance, and reproducibility by others. The last two builds are ALFRED, an aluminium strut-frame robot that collected a year of monthly forest surveys, and ALFRED~2.0, which adds a second frame layer, a second computer and real-time kinematic (RTK) positioning for autonomous crop-row operation. Since the authors had no access to farmland, forestry served as a proxy, testing the platform's ability to capture seasonal and structural change in living plants over a full annual cycle. For the same reason, ALFRED~2.0's crop-row operation used artificial plants set out in rows in a field.

The contributions of this article are:
\begin{itemize}
    \item the presentation of a modular platform used in three experiments in the course of a year all requiring significantly different builds while not missing scheduled data collection in a year long experiment
    \item a requirement-traced account of the four builds, recording the failure behind each design change
    \item a model-based comparison of the last three builds, in which the usable share of the arm's reachable workspace rises from 34.0\% to 60.0\% and 66.1\%, and the visibility of the frame-mounted LiDAR is compared across build, task, and arm posture
    \item an open release of the design files, robot description, and containerised software
\end{itemize}

The remainder of this article is structured as follows. Section~\ref{sec:alfred_background} reviews data assimilation into crop models, mobile manipulators for repeated crop sensing, and the use of reachability and visibility analysis as design tools. Section~\ref{sec:alfred_robots} introduces ALFRED and ALFRED~2.0 together with the ForestYear3D and ROWCAST experiments they were built for, and Section~\ref{sec:alfred_system_design} describes the hardware and software shared by both builds and the differences between them. Section~\ref{sec:alfred_requirements} states the six design requirements and Section~\ref{sec:alfred_evolution} introduces the four prototypes. Sections~\ref{sec:alfred_r1} to~\ref{sec:alfred_reproducibility} then assess every build against each requirement in turn, including the model-based comparison of manipulator reach and sensor visibility in Section~\ref{sec:alfred_reachability}. Section~\ref{sec:alfred_discussion} sets out the limitations and future work, and Section~\ref{sec:alfred_conclusion} concludes.

\section{Background}
\label{sec:alfred_background}

\subsection{Data assimilation into crop models}
\label{subsec:bg_platforms}

Assimilating in-season observations can reduce the error between crop model estimates and in-field observations \citep{dlamini2023assimReview}. Assimilating leaf area index into WOFOST, \citet{curnel2011osseLai} found that accurate weekly observations throughout the season are required to halve yield error. Their experiments used synthetic data, because assessing an assimilation scheme against real observations requires more observations than can be collected. However, the weight of each observation is not uniform throughout the season. Assimilating a single Sentinel-2 leaf area index estimate from tillering, or from booting to flowering, matched or bettered the biomass simulated from assimilating several estimates taken during stem elongation \citep{tewes2020timingFrequency}. These phenological windows are short, and their dates are unknown in advance. On calibrated data, \cite{wallach2021phenology} tested the ability of 27 modelling groups to predict phenology in wheat, the best model had a mean absolute error of around three days. An error of a few days in the date of a window that is itself only days wide means the informative observation cannot be scheduled reliably in advance. Capturing it requires sampling densely enough around the expected stage to bracket that uncertainty, rather than relying on a single well-timed acquisition. Satellite records cannot guarantee this, because revisit is fixed and acquisitions are lost to cloud \citep{whitcraft2015Cloud}, and the accuracy requirement compounds the difficulty. A ground robot answers both: it can be sent out on the days that matter rather than when an orbit allows, and it observes the canopy at close range and from viewpoints that platforms above the crop cannot reach.

\subsection{Mobile manipulators for long-term crop sensing}
\label{subsec:bg_platforms}

Ground robots can carry heavier sensors and provide more viewing angles than aerial platforms when measuring traits at the plant and organ level \citep{xu2022GroundRobots}. These measurements are non-destructively and repeatable at several time points through a growing season \citep{atefi2021RoboticPhenotyping}. \cite{xu2022GroundRobots} state that a challenge for phenotyping robots is they are often designed for one crop and field layout and do not generalise to other crops or layouts. Even within the same crop and layout perception systems can suffer. \cite{burusa2024attentionNBV} claim that perception systems in greenhouses generalise poorly to variation and struggle with issues such as occlusion. They address this issue through the use of a manipulator and an end-effector mounted RGB-D camera deployed upon a cart which could be manoeuvred around the greenhouse.

Repeated observation of the same vegetation has been demonstrated over many months, with sensors carried by hand \citep{knights2023WildPlaces}, backpack \citep{malladi2025DigiForests} and on ground robots carrying fixed sensors \citep{bender2020Ladybird, polvara2024BLT, boxan2026FoMo}. The demands placed on these robots range from battery problems cutting deployments short \citep{boxan2026FoMo} to map reuse degradation in vineyards as the crop grows \citep{polvara2024BLT}. To the best of the authors' knowledge, however, no in-field campaign has monitored the same plants across a full growing season using a mobile manipulator, and none reports the design requirements the platform was built against or releases the information needed to rebuild it.

\subsection{Reachability and visibility as design tools}
\label{subsec:bg_design}

When designing a mobile manipulator for in-field plant monitoring it is useful to quantify the sensor coverage a build provides. Coverage here has two components: the set of poses at which the manipulator can place a sensor, which is a question of reachability, and what the frame-mounted sensors can see from the vehicle, which is a question of visibility. Both can be computed from a model of the robot before it is built, and the tools for doing so are established elsewhere.

Reachability is commonly represented as a capability map, which records where and in which orientations an arm can reach \citep{zacharias2007CapabilityMap}, and inverting such a map yields the base positions from which a target is reachable \citep{vahrenkamp2013ReachabilityInversion}. Agriculture applies this at the design stage with \cite{vanHenten2009OptimalManipulator} selecting a manipulator's kinematic structure by evaluating candidate configurations against the collision-free workspace the crop demands, and \cite{bloch2018OrchardArchitecture} co-designed orchard architecture together with the robot. In both cases the analysis was performed to test the ability of the arm to interact with the environment, rather than evaluate sensor coverage. 

Visibility is treated in the same spirit in autonomous driving, where the placement of one or more LiDAR units is optimised against sensing coverage \citep{kim2020LidarPlacement} and against downstream object detection \citep{hu2022MultiLidarPlacement}. Since the carrier is a rigid body, self-occlusion is identical in every scan. This however does not hold for the mobile manipulator where the largest occluder is the arm that moves independently of the vehicle. Therefore the field of view depends on posture as well as on mounting position. The two components are also coupled, because the structure mounted around the arm that limits its reach is largely the same structure that blocks the vehicle's own LiDAR, so a change intended to improve one alters the other.

This paper reports one mobile manipulator developed through four builds against six requirements, deployed monthly for a year and rebuilt for autonomous crop-row operation while that year was still running. Reachability and visibility are computed on the robot descriptions to compare successive builds before anything is manufactured, and the design files, robot descriptions and containerised software are released.

\section{The ALFRED robots}
\label{sec:alfred_robots}

ALFRED was designed to collect the data needed to improve crop model predictions through data assimilation over multiple growing seasons. The platform is modular because it is shared between several researchers and must change as experimental requirements change, so components need to be added or removed quickly. A mobile manipulator was chosen because sensor placement must change as plants grow, and improving crop model predictions is likely to require several sensors, each with its own placement requirements. Two versions of the platform have been deployed: ALFRED in a year-long forest campaign and ALFRED~2.0 in crop-row trials.

\begin{figure}[htbp]
    \centering
    \begin{subfigure}[b]{0.3\linewidth}
        \centering
        \includegraphics[width=\linewidth]{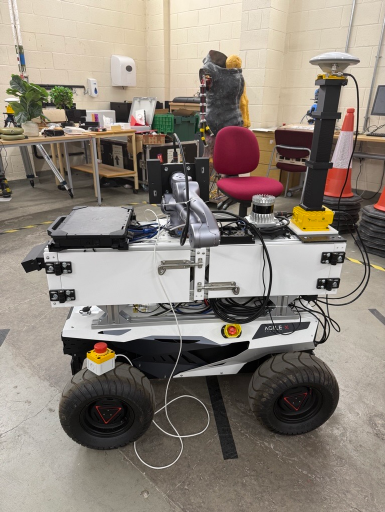}
        \caption{Doors closed}
        \label{fig:alfred_doors_closed}
    \end{subfigure}
    \hfill
    \begin{subfigure}[b]{0.3\linewidth}
        \centering
        \includegraphics[width=\linewidth]{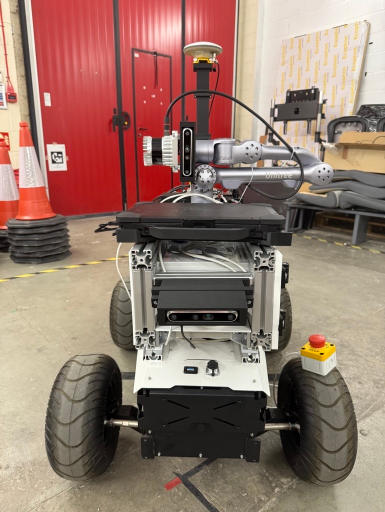}
        \caption{Front view}
        \label{fig:alfred_front_view}
    \end{subfigure}
    \hfill
    \begin{subfigure}[b]{0.3\linewidth}
        \centering
        \includegraphics[width=\linewidth]{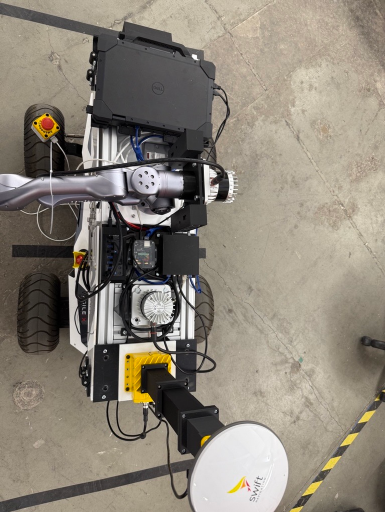}
        \caption{Top-down view}
        \label{fig:alfred_top_down}
    \end{subfigure}
    \caption{ALFRED as configured for the ForestYear3D campaign, showing the aluminium strut frame, hinged access panels, manipulator and sensor suite.}
    \label{fig:alfred_views}
\end{figure}

\begin{figure}[htbp]
    \centering
    \begin{subfigure}[t]{0.45\linewidth}
        \centering
        \includegraphics[width=\linewidth]{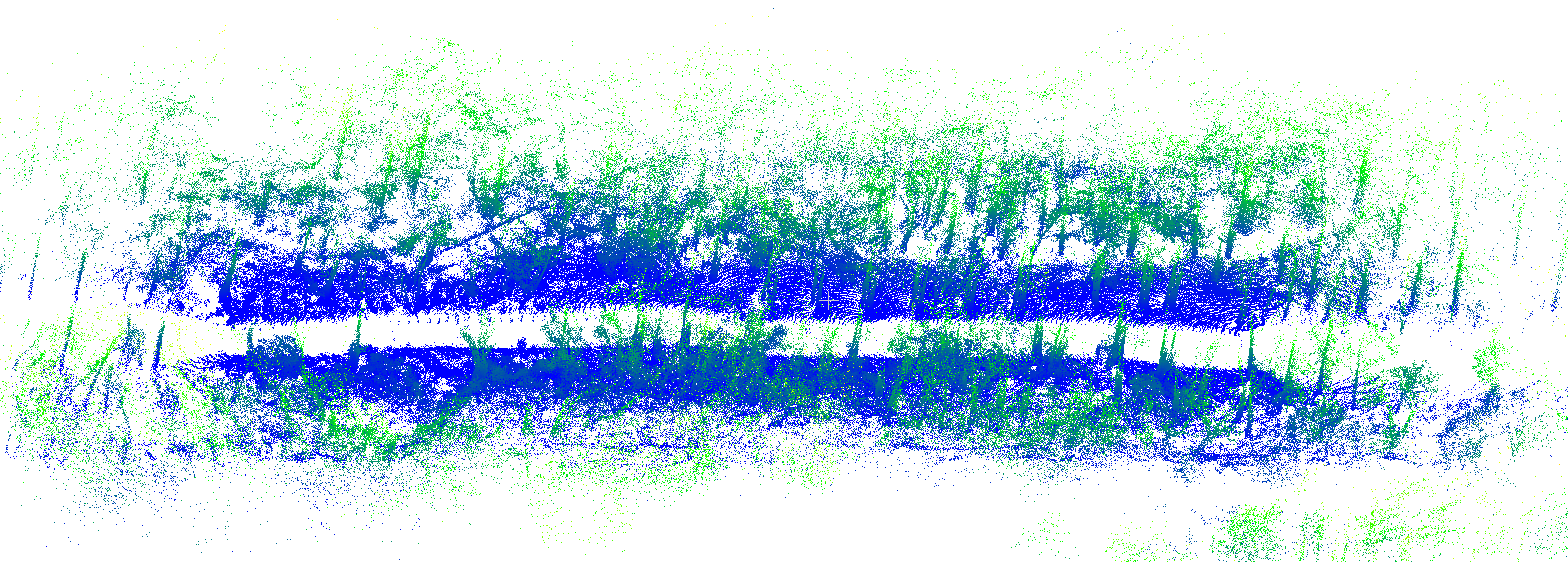}
        \caption{Frame-mounted LiDAR, 2 June 2025}
        \label{fig:base_pcd}
    \end{subfigure}
    \hfill
    \begin{subfigure}[t]{0.45\linewidth}
        \centering
        \includegraphics[width=\linewidth]{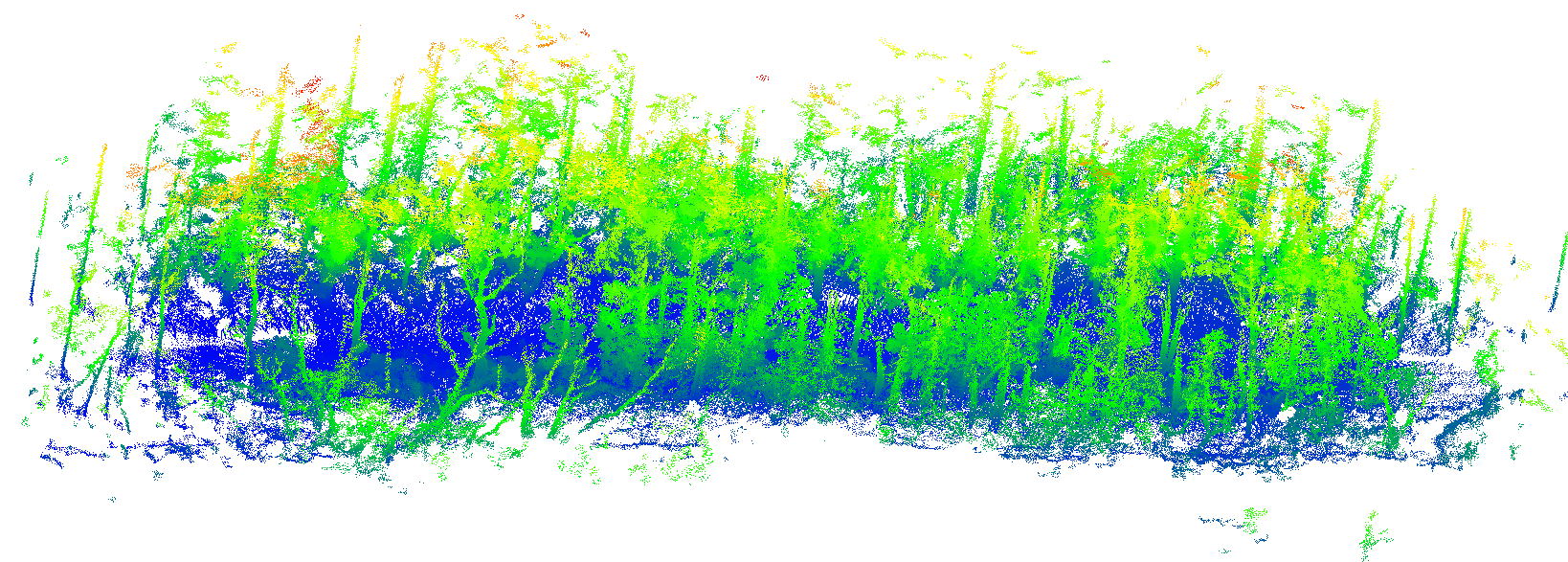}
        \caption{Arm-mounted LiDAR, 2 June 2025}
        \label{fig:arm_pcd}
    \end{subfigure}
    \par\medskip
    \begin{subfigure}[t]{0.45\linewidth}
        \centering
        \includegraphics[width=\linewidth]{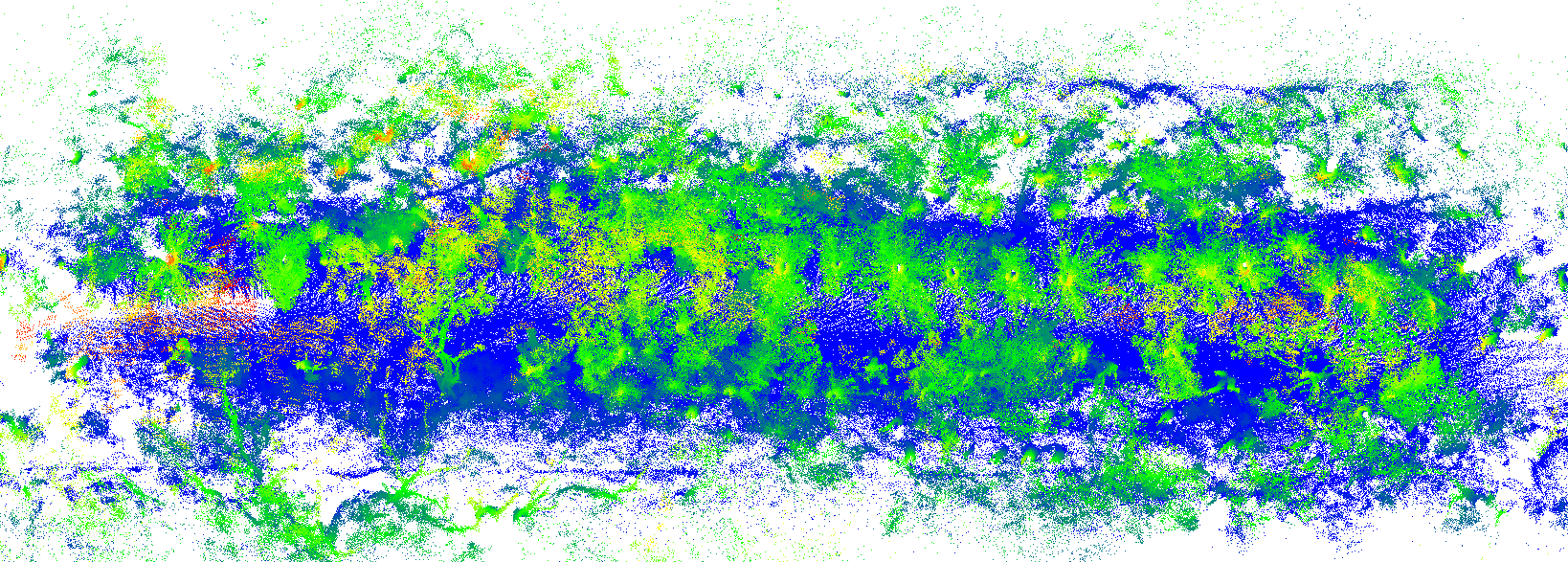}
        \caption{Merged, 2 June 2025}
        \label{fig:summer}
    \end{subfigure}
    \hfill
    \begin{subfigure}[t]{0.45\linewidth}
        \centering
        \includegraphics[width=\linewidth]{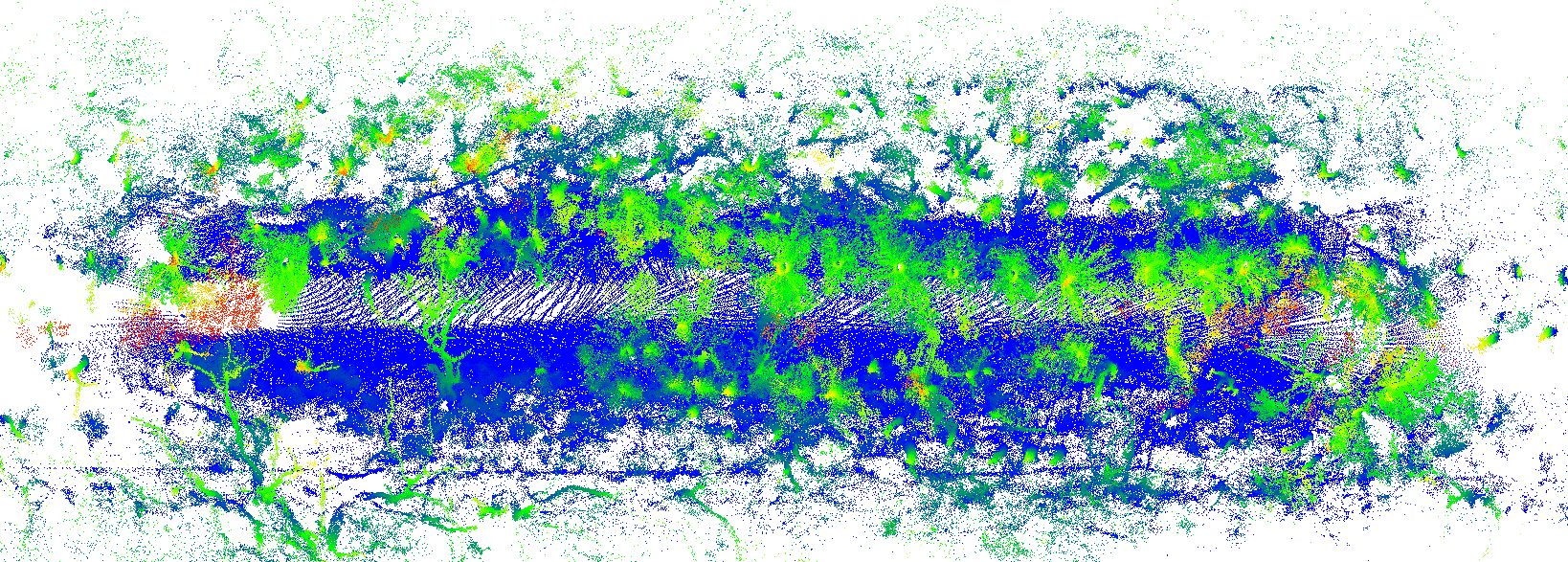}
        \caption{Merged, 9 February 2026}
        \label{fig:winter}
    \end{subfigure}
    \caption{Point clouds of the same forest corridor from ForestYear3D, showing the contribution of each LiDAR and the change between June and February.}
    \label{fig:forestyear3d_data}
\end{figure}

ALFRED (Figure~\ref{fig:alfred_views}) collected the ForestYear3D dataset in the forest next to Heriot-Watt University's Edinburgh campus between June 2025 and May 2026, with the forest serving as a proxy for farmland. Over 29 days, 528 traversals produced 31.68 hours and 4.9 TB of raw ROS bag data. Figure~\ref{fig:forestyear3d_data} shows example point clouds from the campaign. Comparing the frame- and arm-mounted LiDAR clouds from the same day (Figures~\ref{fig:base_pcd} and~\ref{fig:arm_pcd}) shows the additional viewpoints the arm provides. Comparing the merged clouds from June and February (Figures~\ref{fig:summer} and~\ref{fig:winter}) shows the seasonal change in forest structure that the platform captures. Further details are available on the project \href{https://ciaranjohnson.github.io/ForestYear3D/}{\underline{website}} and \href{https://github.com/CiaranJohnson/ForestYear3D}{\underline{GitHub repository}}.

\begin{figure}[htbp]
    \centering
    \begin{subfigure}[b]{0.35\linewidth}
        \centering
        \includegraphics[width=\linewidth]{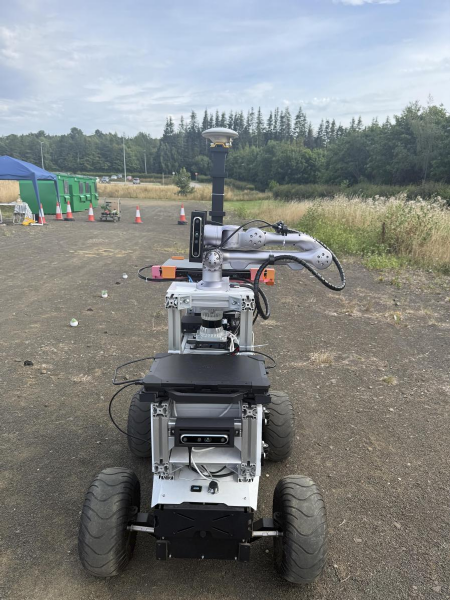}
        \caption{Front view}
        \label{fig:alfred2_front}
    \end{subfigure}
    \hspace{1cm}
    \begin{subfigure}[b]{0.35\linewidth}
        \centering
        \includegraphics[width=\linewidth]{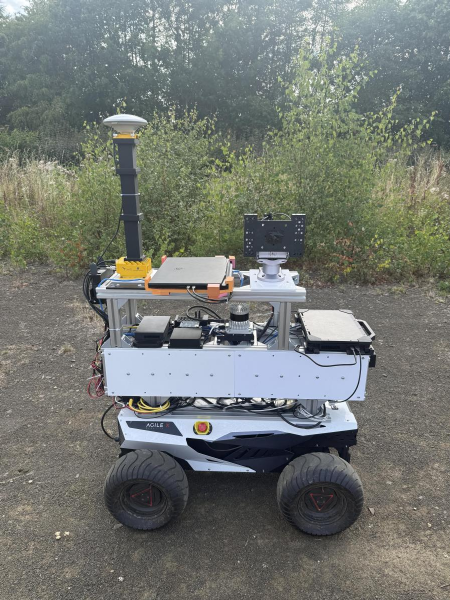}
        \caption{Side view}
        \label{fig:alfred2_side}
    \end{subfigure}
    \caption{ALFRED~2.0 as configured for the ROWCAST experiments, showing the second frame layer, the second laptop and the removed LiDAR.}
    \label{fig:alfred2_views}
\end{figure}

\begin{figure}[htbp]
    \centering
    \begin{subfigure}[t]{0.29\columnwidth}
        \centering
        \includegraphics[width=\linewidth]{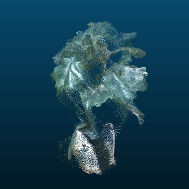}
        \caption{Fiddle-leaf fig}
        \label{fig:fig_plant}
    \end{subfigure}
    \hspace{1cm}
    \begin{subfigure}[t]{0.32\columnwidth}
        \centering
        \includegraphics[width=\linewidth]{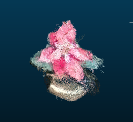}
        \caption{Poinsettia}
        \label{fig:poinsettia}
    \end{subfigure}
    \caption{RGB-D point clouds of two artificial plants reconstructed by ALFRED~2.0 during the ROWCAST trials.}
    \label{fig:rowcast_data}
\end{figure}

ALFRED~2.0 (Figure~\ref{fig:alfred2_views}) was used in the ROWCAST experiments, which automate the collection of row-crop phenotyping data. ROWCAST exploits the semi-static layout of crop rows. A one-off commissioning step establishes the row layout and planting density and uses a costmap to generate a base path that avoids the crops. At each measurement date, the robot then runs a scout-plan-execute-phenotype loop: a fast LiDAR-only scout builds an occupancy map, RGB-D camera viewpoints are planned by ray casting against this map, the plan is executed in the field, and phenotypes are extracted from the collected data. Figure~\ref{fig:rowcast_data} shows RGB-D point clouds of two artificial plants, which are of sufficient quality to extract geometric traits such as plant height, width and canopy cover. Further details are available on the project \href{https://ciaranjohnson.github.io/ROWCAST/}{\underline{website}}.

\section{System Design}
\label{sec:alfred_system_design}
ALFRED and ALFRED~2.0 share almost all of their hardware: an AgileX Hunter 2.0 mobile base~\footnote{\href{https://cdn.shopify.com/s/files/1/0551/0630/6141/files/HUNTER2.0\_USER\_MANUAL.pdf?v=1773112703}{AgileX Hunter 2.0 user manual}}, a Unitree Z1 arm~\footnote{\href{https://www.unitree.com/z1}{Unitree Z1 webpage}\label{fn:z1_web}} and a Dell Latitude 7330 Rugged Extreme laptop~\footnote{\href{https://www.dell.com/en-gb/shop/laptops-2-in-1-pcs/spd/latitude-13-7330-laptop}{Dell Latitude 7330 Rugged Extreme webpage}}, with manufacturer specifications cited for each component.

The Hunter 2.0 is an Ackermann-steered base with two rear-axle drive motors and a third motor for steering. It has a maximum unloaded speed of \SI{1.5}{\metre\per\second}, a maximum steering angle of \SI{33}{\degree} and a maximum vertical payload of \SI{150}{\kilo\gram}. The Z1 is a \SI{4.1}{\kilo\gram} six-joint revolute arm with a maximum reach of \SI{740}{\milli\metre} and a rated end-effector payload of \SI{3}{\kilo\gram}; its joint limits are given in Table~\ref{tab:alfred_joint_limits}. The laptop has an 11th-generation Intel Core i7-1185G7, integrated Intel Xe graphics, \SI{16}{\giga\byte} of memory and a \SI{1}{\tera\byte} solid-state drive, runs Ubuntu 20.04 and hosts the ROS master on both builds.

\begin{table}[htbp]
    \centering
    \caption{Manipulator joint ranges, as specified in the vendor robot description~\footref{fn:z1_web}.}
    \label{tab:alfred_joint_limits}
    \small
    \begin{tabular}{@{}lrr@{}}
        \toprule
        Joint & Minimum [rad] & Maximum [rad] \\
        \midrule
        1 (waist)    & $-2.618$ & $2.618$ \\
        2 (shoulder) & $0$      & $2.967$ \\
        3 (elbow)    & $-2.880$ & $0$ \\
        4 (wrist 1)  & $-1.518$ & $1.518$ \\
        5 (wrist 2)  & $-1.344$ & $1.344$ \\
        6 (wrist 3)  & $-2.793$ & $2.793$ \\
        \bottomrule
    \end{tabular}
\end{table}

The full sensor suite comprises two Ouster OS1-32 LiDARs~\footnote{\href{https://data.ouster.io/downloads/datasheets/datasheet-rev7-v3p1-os1.pdf}{Ouster OS1-32 datasheet}}, one on the arm's end effector and one fixed to the base; an InertialSense RUG-3-IMX-5-RTK inertial unit~\footnote{\href{https://docs.inertialsense.com/datasheets/IMX-5\_IMU\_AHRS\_GNSS-INS\_Datasheet.pdf}{ InertialSense RUG-3-IMX-5-RTK inertial unit datasheet}} beneath the base LiDAR; three Intel RealSense D457 RGB-D cameras~\footnote{\href{https://www.intel.com/content/www/us/en/products/sku/230571/intel-realsense-depth-camera-d457/specifications.html}{Intel RealSense D457 RGB-D webpage}}, one facing forwards on the base and two on the arm; and a SwiftNav Duro RTK GNSS receiver~\footnote{\href{https://www.swiftnav.com/resource-files/Duro/v2.0/Manual/Duro\%20User\%20Manual\%20v2.0\%20[00533-02-0].pdf}{SwiftNav Duro RTK GNSS user manual download}}, raised clear of the frame on a printed mast to reduce multipath and near-field interference from the surrounding metal. To limit data volume, only a subset of sensors is recorded in each experiment. ALFRED carries all sensors but records only the two LiDARs and the IMU. For the ROWCAST experiments, ALFRED~2.0 has the arm-mounted LiDAR removed and records every remaining sensor except the forward-facing camera.

\begin{figure}[t]
    \centering
    \begin{subfigure}[t]{0.32\linewidth}
        \centering
        \includegraphics[width=\linewidth]{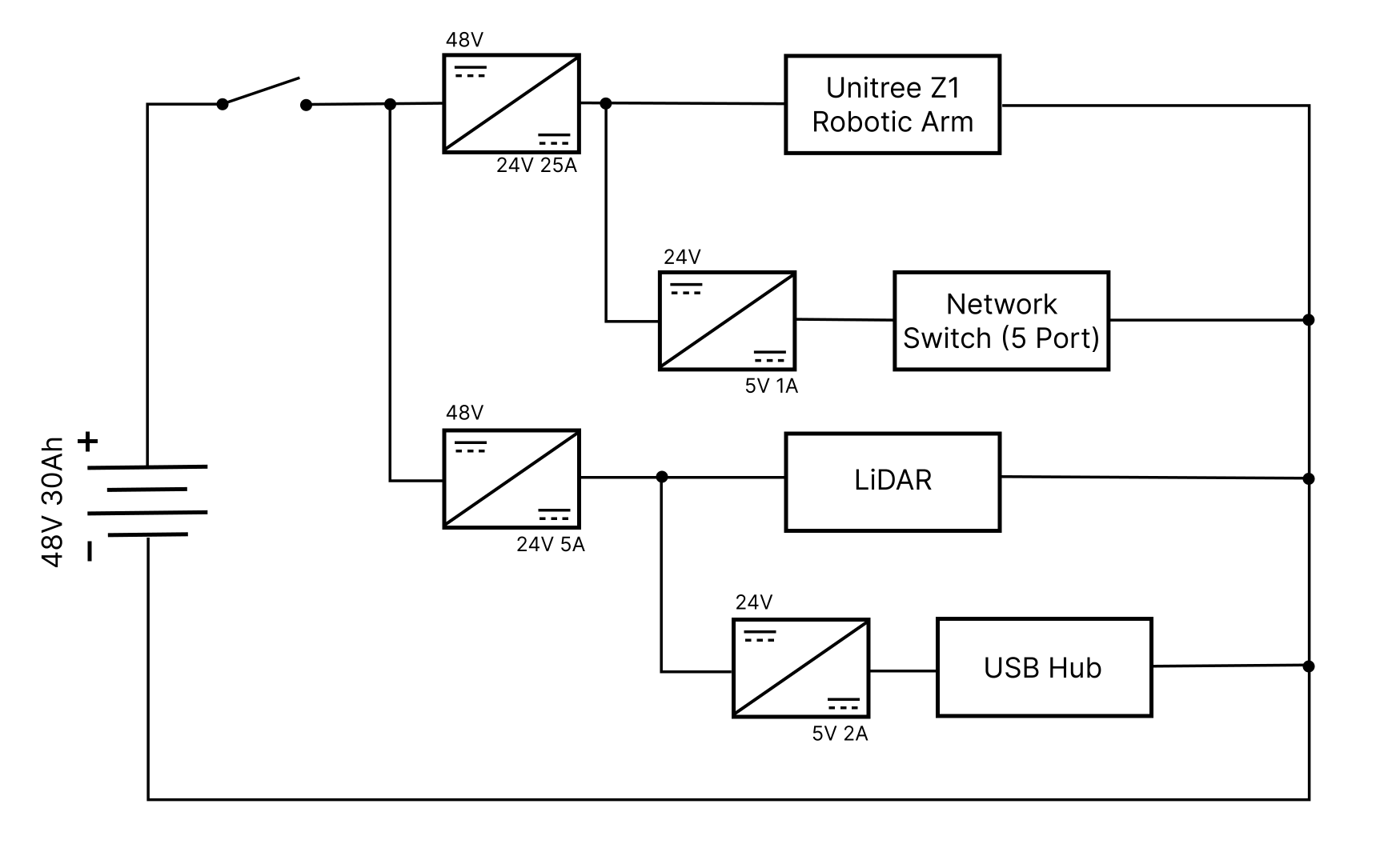}
        \caption{Arm enclosure (both builds).}
        \label{fig:circuit_arm}
    \end{subfigure}
    \hfill
    \begin{subfigure}[t]{0.32\linewidth}
        \centering
        \includegraphics[width=\linewidth]{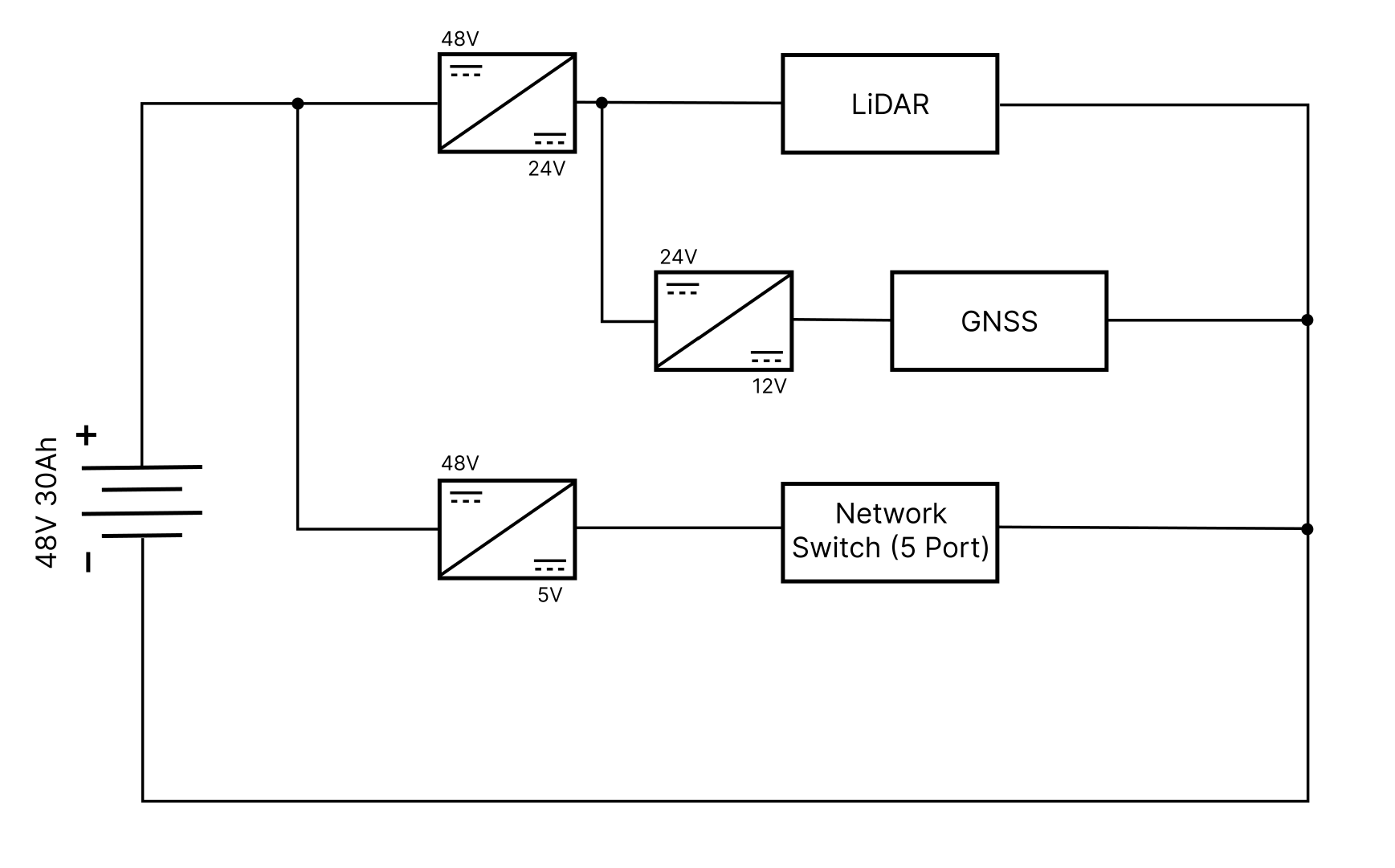}
        \caption{Sensor enclosure, ALFRED.}
        \label{fig:circuit_1}
    \end{subfigure}
    \hfill
    \begin{subfigure}[t]{0.32\linewidth}
        \centering
        \includegraphics[width=\linewidth]{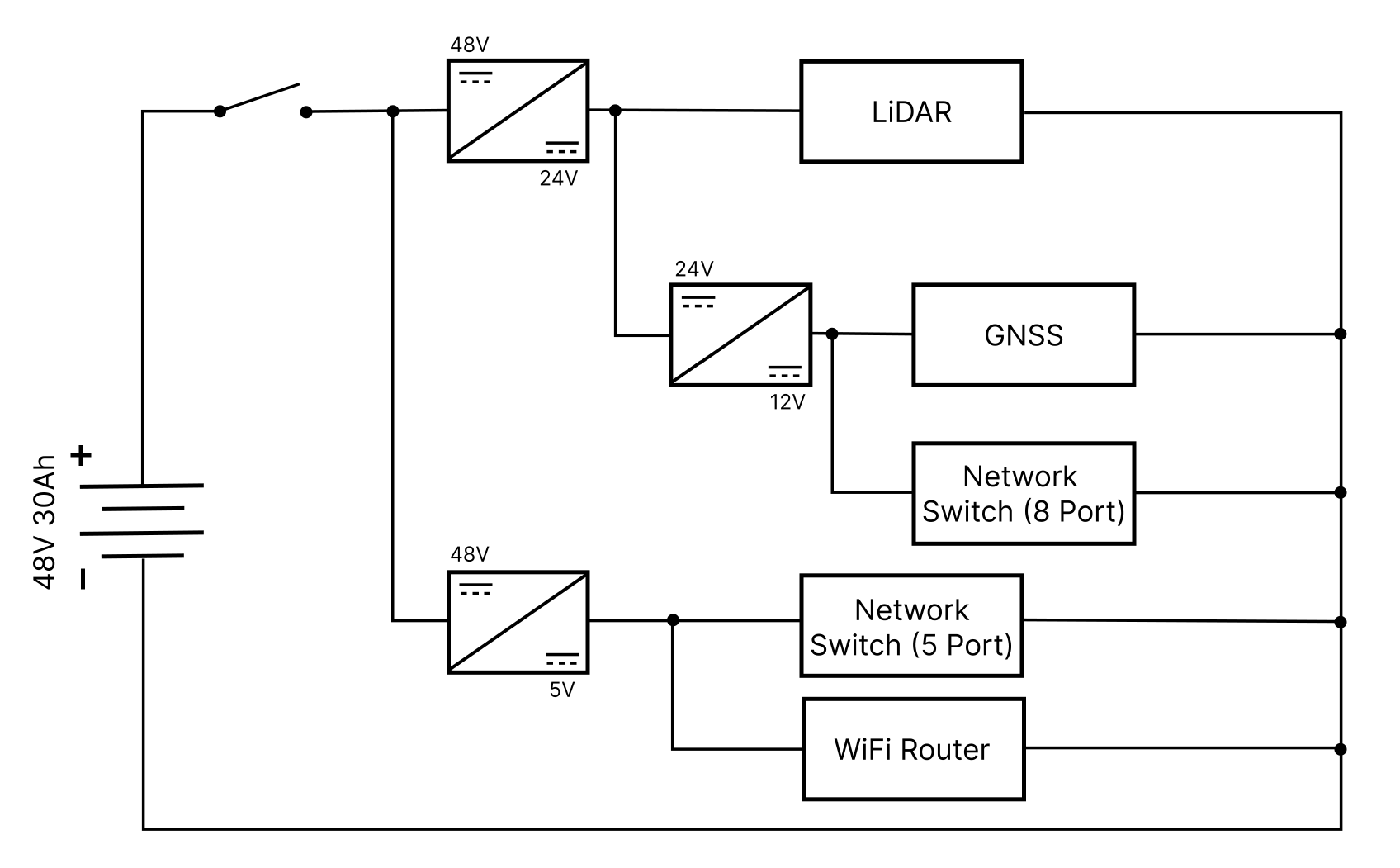}
        \caption{Sensor enclosure, ALFRED~2.0.}
        \label{fig:circuit_2}
    \end{subfigure}
    \caption{Simplified power distribution within the electronics enclosures. Each enclosure has an independent \SI{48}{\volt} \SI{30}{\ampere\hour} battery and steps down to the voltages its loads require.}
    \label{fig:circuit_diagrams}
\end{figure}

The system is powered by two waterproof electronics enclosures, each with its own \SI{48}{\volt} \SI{30}{\ampere\hour} battery and a step-down tree supplying the voltages its loads require, so either can be built, debugged or replaced without affecting the other (Figure~\ref{fig:circuit_diagrams}). The arm enclosure is the same on both builds and powers the arm, the arm-mounted LiDAR, a USB hub and a network switch (Figure~\ref{fig:circuit_arm}). The sensor enclosure differs between builds. On ALFRED it powers the base LiDAR, the GNSS receiver and a five-port network switch (Figure~\ref{fig:circuit_1}). ALFRED~2.0 adds a WiFi router and an eight-port network switch, as well as an external power switch matching the one on the arm enclosure (Figure~\ref{fig:circuit_2}).

Full hardware, electronics and strut-frame component lists and costs, are given in Tables~\ref{tab:hardware_table}, \ref{tab:electronics_table} and \ref{tab:strut_frame_table}.

\subsection{Software}

Both builds run ROS~1 Noetic in Docker containers with a two-layer architecture: a base image holds the core robot configuration, drivers and dependencies, and application images built on top hold the planning and behaviour components~\cite{varpe2025containers}. Containerisation was adopted for reproducibility, portability and environment isolation, since the control laptop was shared with another researcher~\cite{kadri2022containers, varpe2025containers}. The layered structure also shortens build times and lets experimental behaviours be developed without altering the core system.

Wheel odometry drifts under wheel slip and on uneven terrain, so both campaigns build their maps with FAST-LIO SLAM~\cite{}, which combines FAST-LIO2 LiDAR-inertial odometry~\cite{xu2021fastlio2fastdirectlidarinertial} with a Scan Context pose-graph optimisation back end~\cite{}.

The Z1 is controlled through the vendor's ROS wrapper around its SDK~\cite{}, with joint trajectories sent directly to the controller's action server. In the forest campaign these were parametric sweeps between fixed joint configurations. In crop-row operation they were generated and collision-checked offline in Pinocchio~\cite{carpentier2019Pinocchio}, using the semantic robot description and self-collision matrix produced by the MoveIt Setup Assistant~\cite{}. No motion reported in this work was planned online.

Three visualisation and simulation tools are used for different purposes. Gazebo~\cite{} provides rigid-body physics for testing control and navigation. RViz~\cite{} displays live or recorded sensor streams, the transform tree, costmaps and the robot model, serving both as the operator display during sessions and for diagnosing behaviour afterwards. MeshCat~\cite{} provides a lightweight kinematic view without physics or controllers, which makes it the fastest way to iterate on planned motions and to confirm that sensor frame orientations match between the robot and the planner.

\subsection{ALFRED: ForestYear3D configuration}

During the ForestYear3D campaign, ALFRED carried both LiDARs, with the frame-mounted unit operating at $1024\times\SI{10}{\hertz}$ and the end-effector unit at $2048\times\SI{10}{\hertz}$. The \SI{1.590}{\kilo\gram} end-effector payload comprised the manipulator interface, the sensor mount, the arm-mounted LiDAR with its base and radial heatsinks, and two RGB-D cameras. Inertial data was recorded from the InertialSense unit, at X for calibrated readings and 1kHz for raw readings, and from the internal IMUs of both LiDARs. The GNSS receiver was carried but not used, because RTK had not been commissioned by the last date on which a full annual cycle of collection could begin; this was a low priority, as forest corridors are GNSS-degraded environments~\cite{}. Each traversal was recorded as a ROS bag containing the sensor, robot-state, control and transform streams needed to replay the acquisition and rerun the reconstruction.

Operation was semi-autonomous. The manipulator autonomously swept between fixed joint configurations, while the base moved at a constant commanded velocity and an operator following the robot steered it. This removed any dependence on SLAM performing robustly across the seasonal and structural changes in the forest.

\subsection{ALFRED~2.0: ROWCAST configuration}
\label{subsec:alfred_p3}
ALFRED~2.0 was built for autonomous crop-row monitoring. A second onboard laptop and a WiFi router were added, which required a second frame layer and a larger network switch. The arm-mounted LiDAR was removed to reduce data volume and to cut the end-effector payload to \SI{0.530}{\kilo\gram}. The two D457 cameras on opposite sides of the arm payload became the primary sensors, recording at $1280\times720$ or $848\times480$ in separate experiments, while the frame-mounted LiDAR ran at $2048\times\SI{10}{\hertz}$ throughout. The base navigates autonomously along a pre-validated route, using Hybrid~A* with a Timed Elastic Band local planner and a semantic pure-pursuit controller. As on ALFRED, manipulator motions are joint trajectories planned offline and checked for self-collision before deployment.

The second laptop, an MSI Stealth GS77 12UGS (Intel Core i7-12700H, \SI{32}{\giga\byte} memory, \SI{1}{\tera\byte} disk, NVIDIA GeForce RTX 3070 Ti), was added for three reasons. First, the primary laptop lacked the free ports to run the two arm-mounted cameras. Second, ALFRED~2.0 was intended to plan online, and this computation was to be kept separate from the control loop. Third, its GPU would make onboard semantic and instance segmentation feasible. Only the first was realised: the online planner was not brought into service in the time available (Section~\ref{subsec:future_work}), and no learning-based perception ran onboard.

The WiFi router enables RTK positioning, since a standalone GNSS fix is accurate only to metres, which is coarse relative to the scale of a crop row. A dedicated radio link is the conventional and more robust choice for RTK corrections, but WiFi was adopted because it serves a second purpose at no extra hardware cost: it makes the onboard computers reachable from an operator laptop at the field edge, so a session can be started, monitored and visualised remotely. The operator laptop is an Alienware m16 R1 (Intel Core i9-13900HX, \SI{64}{\giga\byte} memory, \SI{4}{\tera\byte} disk). All three laptops run their robot software in ROS Noetic Docker containers.

\begin{figure}[htbp]
    \centering
    \includegraphics[width=0.65\linewidth]{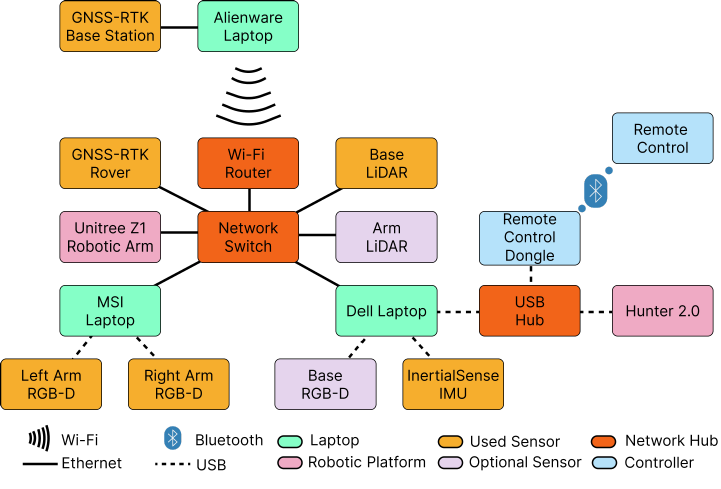}
    \caption{Network topology of ALFRED~2.0. Colour distinguishes computers, network hubs, control hardware, and sensors used in the reported experiments from those carried but not used. Line style distinguishes Ethernet, USB, Wi-Fi and Bluetooth connections. The wireless link carries RTK correction messages from the base station to the rover.}
    \label{fig:network_config}
\end{figure}

With three computers on wired and wireless links, every interface was given a fixed address. The Dell laptop remains the ROS master because it carries the safety-critical interfaces. Figure~\ref{fig:network_config} shows the network topology.

\section{Design requirements}
\label{sec:alfred_requirements}

\begin{figure}[t]
    \centering
    \begin{subfigure}[t]{0.24\columnwidth}
        \centering
        \includegraphics[width=\linewidth]{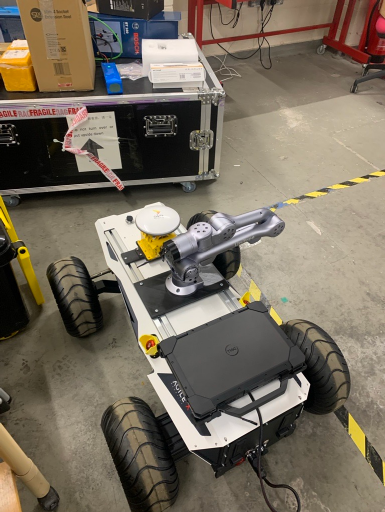}
        \caption{Prototype~0}
        \label{fig:alfred_p0}
    \end{subfigure}
    \hfill
    \begin{subfigure}[t]{0.24\columnwidth}
        \centering
        \includegraphics[width=\linewidth]{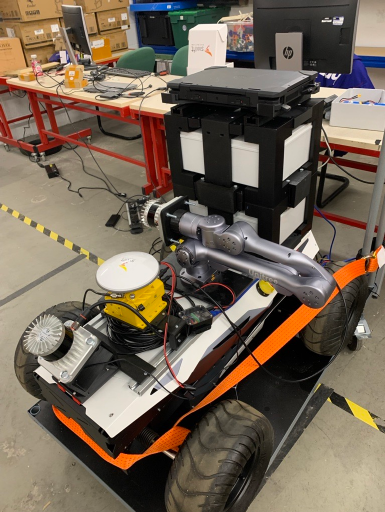}
        \caption{Prototype~1}
        \label{fig:alfred_p1}
    \end{subfigure}
    \hfill
    \begin{subfigure}[t]{0.24\columnwidth}
        \centering
        \includegraphics[width=\linewidth]{Figures/prototyping/doors_closed.pdf}
        \caption{ALFRED}
        \label{fig:alfred_p2}
    \end{subfigure}
    \hfill
    \begin{subfigure}[t]{0.243\columnwidth}
        \centering
        \includegraphics[width=\linewidth]{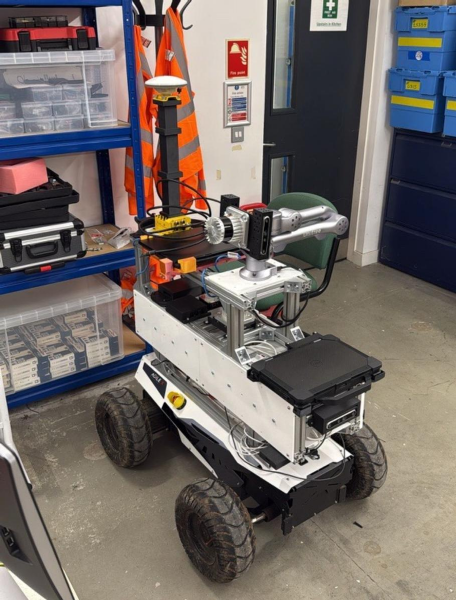}
        \caption{ALFRED~2.0}
        \label{fig:alfred_p3}
    \end{subfigure}
    \caption{The four builds of the platform. Prototype~0 established component integration, Prototype~1 was the first complete assembly, ALFRED collected the ForestYear3D dataset, and ALFRED~2.0 added the computing, networking and sensing required for autonomous crop-row operation.}
    \label{fig:alfred_prototypes}
\end{figure}

Both robots were built iteratively against six design requirements, listed below and followed by a discussion of how they shaped the four builds of the platform (Figure~\ref{fig:alfred_prototypes}).

\paragraph{R1: Outdoor durability} The platform must operate unsheltered across a full annual cycle without loss of function. It must survive unforecast light rain or gusts mid-session, tolerate mud, dust, pollen, humidity and the site's annual temperature range, and traverse fallen leaves, loose stones and exposed roots.

\paragraph{R2: Modularity and reconfigurability} Sensors, computers and other mounted hardware must be relocatable and replaceable without significant redesign, so the platform can be reconfigured as sensing requirements change and shared with other researchers.

\paragraph{R3: Repairability} Failures must be repairable in-house between scheduled deployments, using commercially available components or parts made with available workshop equipment.

\paragraph{R4: Sensing reach and coverage} The platform must place sensors at poses a rigidly mounted sensor cannot reach, and provide enough base-mounted field of view for each task.

\paragraph{R5: Endurance} The platform must complete a scheduled data collection session without being limited by battery capacity.

\paragraph{R6: Reproducibility by others} A third party must be able to rebuild the platform: bespoke components manufacturable from released design files, other components commercially available and priced, and the software stack portable across machines.

\section{Platform development}
\label{sec:alfred_evolution}

This section assesses the four builds shown in Figure~\ref{fig:alfred_prototypes} (Prototype~0, Prototype~1, ALFRED and ALFRED~2.0) against the design requirements, as summarised in Figure~\ref{fig:design_req}. All four are presented as prototypes rather than finished products, since many components could be further optimised and validated.

\begin{figure}[htbp]
    \centering
    \includegraphics[width=0.7\linewidth]{Figures/Additional/design_requirements.pdf}
    \caption{Each of the four builds assessed against the six design requirements as not satisfied, partially satisfied or satisfied.}
    \label{fig:design_req}
\end{figure}

\subsection{Prototype overview}

Prototype~0 was a feasibility build, used to develop the manufacturing skills later builds would need and to confirm that the equipment could be mounted on the Hunter~2.0 base. It is discussed below only where it partially satisfied a requirement.

Prototype~1 was the first build to integrate the main subsystems: the Hunter~2.0 base, the Z1 arm, two OS1-32 LiDARs, the Duro GNSS receiver and a control laptop. Its main advance was independence from external power, through the two independently powered enclosures. It was never deployed in the field: bench assembly and laboratory testing showed that it failed a requirement. From then on, every build was modelled in CAD and tested in simulation before construction, so that such problems were found before anything was manufactured.

\begin{figure}[htbp]
    \centering
    \includegraphics[width=0.35\linewidth]{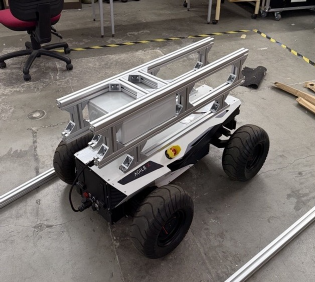}
    \caption{ALFRED's aluminium profile strut frame before panelling and components were mounted.}
    \label{fig:aluminium_frame}
\end{figure}

ALFRED was the first build to meet every requirement at least partially and the first to be used in field experiments. The main change was a modular frame of aluminium profile struts (Figure~\ref{fig:aluminium_frame}).

ALFRED~2.0 addressed the sensor coverage limitations that prevented the move from semi-autonomous operation to autonomous crop-row monitoring, mainly through a second frame layer, a second computer and a WiFi router.

\section{R1: Outdoor durability}
\label{sec:alfred_r1}

Prototype~1 failed R1 because of the 3D-printed cage housing its electronics enclosures (Figure~\ref{fig:alfred_p1}). The cage was printed in PLA at low infill, since higher infill was prohibitively slow to print, and PLA performs poorly under sustained load and degrades with ultraviolet exposure and moisture. Several parts of the cage broke during assembly, and the rest were judged likely to fail in the field. ALFRED replaced the cage with a frame of aluminium profile struts and angle brackets. No structural element of this frame has failed in the two years since it was built. This does not prove durability, but it shows the frame was sufficient for the work performed, and the same struts were therefore used for ALFRED~2.0's second layer. Both ALFRED and ALFRED~2.0 satisfy R1.

\section{R2: Modularity and reconfigurability}
\label{sec:alfred_r2}

\begin{figure}[htbp]
    \centering
    \begin{subfigure}[t]{0.23\linewidth}
        \centering
        \includegraphics[width=\linewidth]{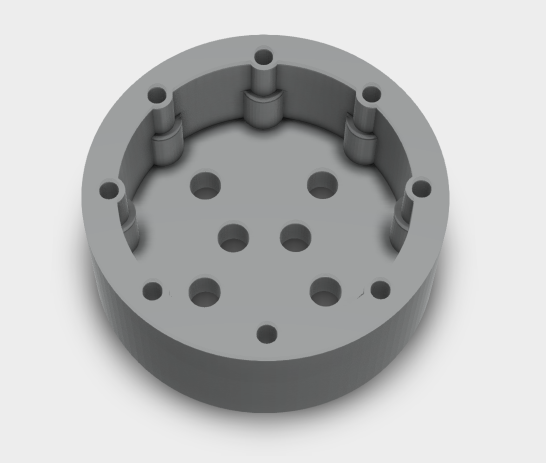}
        \caption{End-effector interface.}
        \label{fig:ee_interface}
    \end{subfigure}
    \hfill
    \begin{subfigure}[t]{0.23\linewidth}
        \centering
        \includegraphics[width=\linewidth]{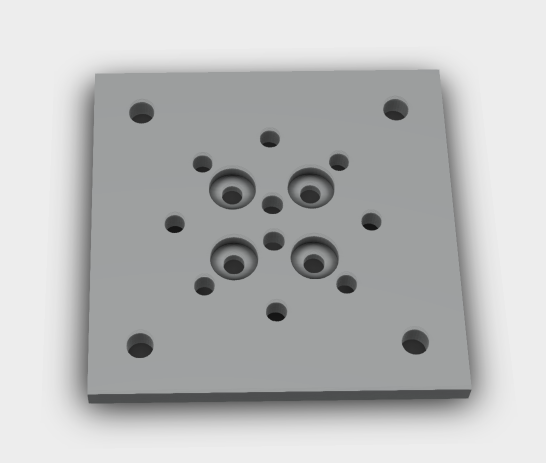}
        \caption{LiDAR end effector.}
        \label{fig:alfred_lidar_effector}
    \end{subfigure}
    \hfill
    \begin{subfigure}[t]{0.23\linewidth}
        \centering
        \includegraphics[width=\linewidth]{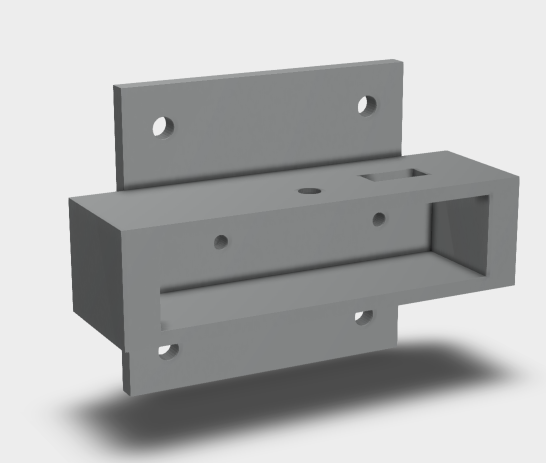}
        \caption{RGB-D end effector.}
        \label{fig:alfred_rgbd_effector}
    \end{subfigure}
    \hfill
    \begin{subfigure}[t]{0.23\linewidth}
        \centering
        \includegraphics[width=\linewidth]{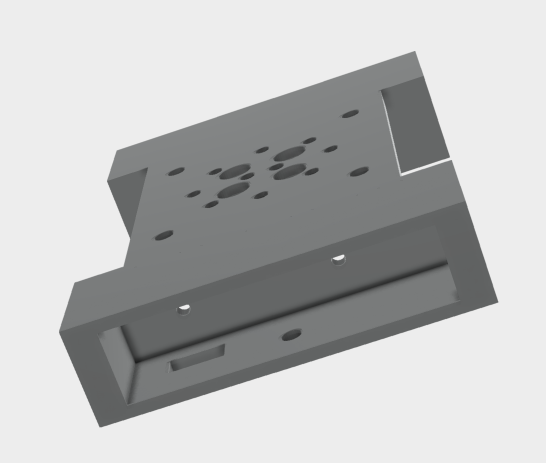}
        \caption{Combined end effector.}
        \label{fig:alfred_combined_effector}
    \end{subfigure}
    \caption{Interchangeable sensor mounts, each built on the common manipulator interface.}
    \label{fig:alfred_sensor_mounts}
\end{figure}

Prototype~0 established the design principles for R2, chief among them a 3D-printed end-effector interface (Figure~\ref{fig:ee_interface}) that carries all the geometry needed to mate with the manipulator flange. Each new payload therefore needed only its own housing to be designed and printed, rather than the whole assembly, saving time and material (Figure~\ref{fig:alfred_sensor_mounts}). Shared interfaces later served the same purpose in software, keeping the codebase smaller and easier to maintain. The interface survived unchanged to the final design, and Prototype~0's components could be rearranged easily, but as a non-functioning build it only partially satisfies R2.

Prototype~1 also only partially satisfies R2, because it lacked mounting space. The exposed profile strut on the Hunter~2.0 offers little area for the number of components the platform carries, so the printed enclosure cages were stacked vertically with the laptop on a printed rail above them. Components could be rearranged only within tight constraints, there was no room for additional hardware, and the enclosures could not be reached in the field without partial disassembly.

\begin{figure}[htbp]
    \centering
    \includegraphics[width=0.35\linewidth]{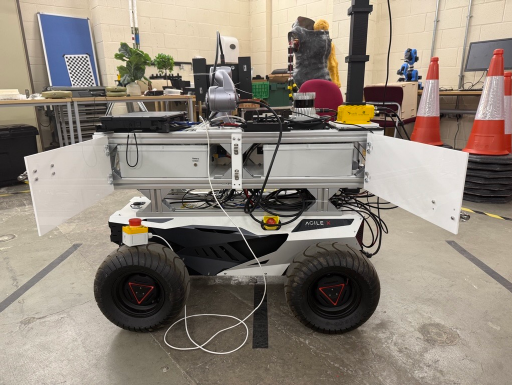}
    \caption{ALFRED with a hinged door open, giving field access to an electronics enclosure.}
    \label{fig:doors_open}
\end{figure}

ALFRED's aluminium frame extended the mounting area beyond the footprint of the base. Laser-cut acrylic panels converted it into two housings for the electronics enclosures, each behind a hinged, latched door that gives access in the field (Figure~\ref{fig:doors_open}). ALFRED~2.0 extended the same strut design with a second layer to carry the additional hardware for the ROWCAST experiments.

\subsection{Reconfiguration during the ForestYear3D campaign}

\begin{figure}[htbp]
    \centering
    \includegraphics[width=0.35\linewidth]{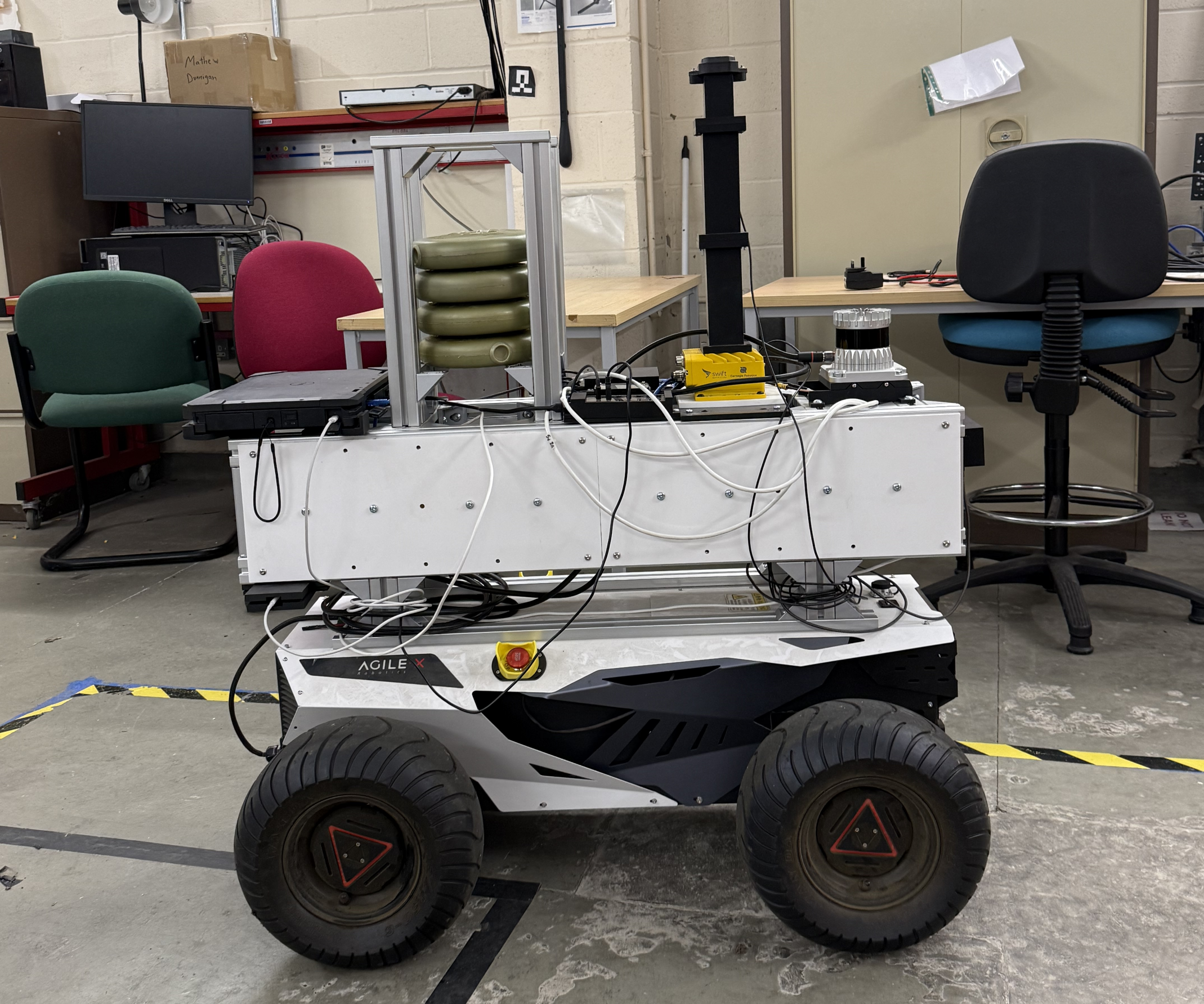}
    \caption{ALFRED reconfigured for the concurrent payload study, with the manipulator replaced by a weight cage and an alternative GNSS receiver fitted.}
    \label{fig:alfred_payload_config}
\end{figure}

The strongest evidence for R2 is that, during a year of monthly forest surveys, ALFRED was repeatedly reconfigured for another researcher's study and for the development of ALFRED~2.0 without missing a scheduled collection. Two months into the campaign, the platform was reconfigured to study predicting added payload mass and classifying manoeuvres from the base's battery-current response. The manipulator was replaced with a cage carrying weights, and an alternative GNSS receiver was fitted using a bespoke adapter (Figure~\ref{fig:alfred_payload_config}). Each reconfiguration took about half a day in each direction.

\begin{figure}[htbp]
    \centering
    \begin{subfigure}[t]{0.42\linewidth}
        \centering
        \includegraphics[width=\linewidth]{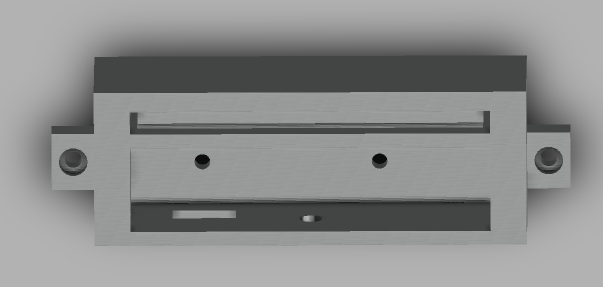}
        \caption{Vented fixed-camera mount.}
        \label{fig:alfred_vented_front}
    \end{subfigure}
    \hspace{1cm}
    \begin{subfigure}[t]{0.24\linewidth}
        \centering
        \includegraphics[width=\linewidth]{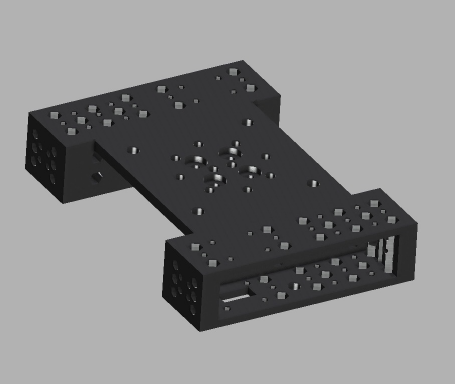}
        \caption{Vented end-effector mount.}
        \label{fig:alfred_vented_effector}
    \end{subfigure}
    \caption{RGB-D mounts redesigned with ventilation after the cameras overheated during extended operation.}
    \label{fig:alfred_vented_mounts}
\end{figure}

This period also exposed an early fault. In August 2025, the printed camera housings were found to restrict airflow enough for the RGB-D cameras to overheat during extended operation. Both the fixed and end-effector mounts were redesigned with ventilation openings (Figure~\ref{fig:alfred_vented_mounts}). The thermal improvement was not quantified, but the fault did not recur, and finding it early meant it was resolved before the ROWCAST experiments.

Because ALFRED~2.0 was developed before the ForestYear3D campaign ended, the platform switched repeatedly between the two builds, and each complete transition took about six hours. These transitions were not purely mechanical. Each build has its own robot description, collision matrix, sensor frames and inertial properties, and a platform launched against the wrong description will run but behave incorrectly, which can be dangerous. The same risk exists within a build: ALFRED~2.0 can be fitted with or without the arm-mounted LiDAR, and the controller must be updated for the change in payload. A mechanically modular platform is therefore only modular if its software model tracks its mechanical state, and this coupling is where modularity leaks. With both hardware and software kept in step through these transitions, ALFRED and ALFRED~2.0 satisfy R2.

\section{R3: Repairability}
\label{sec:alfred_r3}

Prototype~0 partially satisfies R3: every component was made in-house by 3D printing or laser cutting, and the end-effector interface (Section~\ref{sec:alfred_r2}) means a broken mount can be replaced without reprinting the interface.

Prototype~1's power system was designed for repairability. The base can supply power directly, but unregulated, and designing a single integrated distribution system needed electronics experience that was not available at the time. Separate enclosures with their own batteries and step-down converters were adopted instead, trading mass and efficiency for the ability to build, debug and replace each supply independently. This trade is typical of the whole platform: where the optimal solution and the repairable one conflicted, the repairable one was chosen.

\begin{figure}[t]
    \centering
    \begin{subfigure}[t]{0.35\columnwidth}
        \centering
        \includegraphics[width=\linewidth]{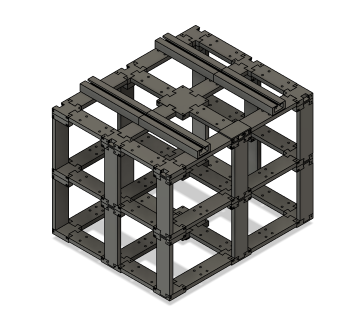}
        \caption{First design.}
        \label{fig:box_p1}
    \end{subfigure}
    \hspace{1cm}
    \begin{subfigure}[t]{0.35\columnwidth}
        \centering
        \includegraphics[width=\linewidth]{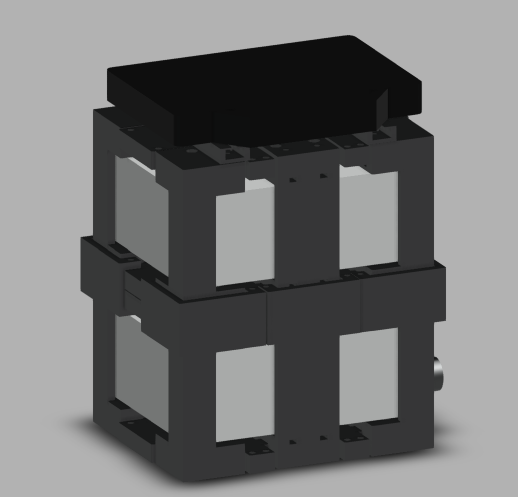}
        \caption{Second design.}
        \label{fig:box_p2}
    \end{subfigure}
    \caption{CAD designs for Prototype~1's 3D-printed cage housing the electronics enclosures.}
    \label{fig:box_prototypes}
\end{figure}

Prototype~1's printed cage, however, meant it only partially satisfied R3. The first design (Figure~\ref{fig:box_p1}) was over-engineered, inspired by Japanese wood joinery for its elegance rather than any property the platform needed, and was slow to model and slower to print. The lesson is one the iterative approach is meant to surface: effort spent perfecting a subsystem before the platform exists is spent against assumptions rather than observed requirements. The simpler second design (Figure~\ref{fig:box_p2}) was printed in PLA at low infill to reduce print times, and, as described in Section~\ref{sec:alfred_r1}, many parts broke during assembly. Although each part could be reprinted in a single working day, this could cost a scheduled collection, particularly in months with two collections, and a shared material weakness makes several parts likely to fail together. The immediate response was to switch from PLA to PETG, which is less brittle and more tolerant of outdoor exposure \citep{sedlak2023DegradationFDM}.The more important conclusion was that the load-bearing structure should not be printed at all.

ALFRED's frame is built from stock materials. The profile strut is bought by the metre and cut to length with a hacksaw, bench vice and file, and the acrylic panels are laser-cut from sheet held in stock, so a damaged strut or panel can be replaced in minutes rather than a working day of printing. Both field builds were repaired without disrupting their schedules. During the ForestYear3D campaign, the combined RGB-D and LiDAR end-effector mount was damaged and replaced the following day, with no effect on data collection. On ALFRED~2.0, a loose power connector intermittently cut supply to the new switch and router; it was replaced with a threaded connector and the fault did not recur. Both ALFRED and ALFRED~2.0 therefore satisfy R3.

\section{R4: Sensing reach and coverage}
\label{sec:alfred_reachability}

\begin{figure}[htbp]
    \centering
    \begin{subfigure}[t]{0.32\columnwidth}
        \centering
        \includegraphics[width=\linewidth]{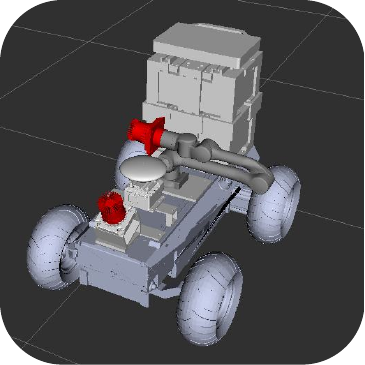}
        \caption{Prototype~1.}
        \label{fig:alfred_sim_p1}
    \end{subfigure}
    \hfill
    \begin{subfigure}[t]{0.32\columnwidth}
        \centering
        \includegraphics[width=\linewidth]{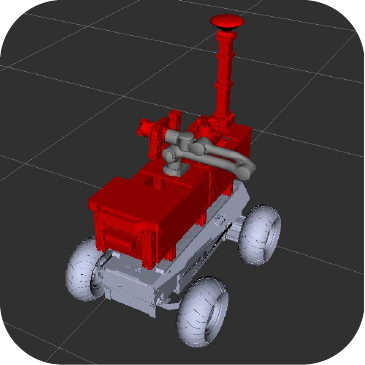}
        \caption{ALFRED.}
        \label{fig:alfred_sim_p2}
    \end{subfigure}
    \hfill
    \begin{subfigure}[t]{0.32\columnwidth}
        \centering
        \includegraphics[width=\linewidth]{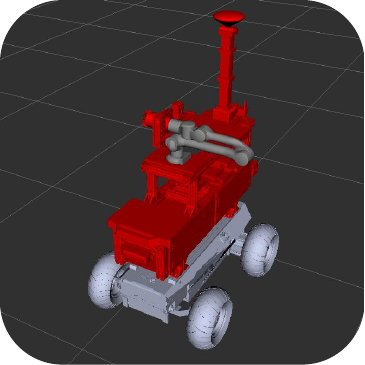}
        \caption{ALFRED~2.0.}
        \label{fig:alfred_sim_p3}
    \end{subfigure}
    \caption{Robot descriptions of the three complete builds, with the manipulator in the zero posture, used for the visibility and reachability analyses.}
    \label{fig:alfred_sim_builds}
\end{figure}

R4 has two parts: the frame-mounted LiDAR must keep enough visibility for each experiment, and the manipulator must be able to place an end-effector-mounted sensor at poses the frame-mounted sensors cannot reach. Each complete build is assessed against both using its robot description (Figure~\ref{fig:alfred_sim_builds}). Meshes for the commercial base, manipulator and sensors come from manufacturer models, and those for the bespoke parts and the aluminium frame were exported from the CAD used to build them. The comparisons therefore test every build under identical assumptions, but they do not characterise field accuracy: cable routing and small fasteners are not modelled, so all three collision models are slightly optimistic.

\subsection{Fixed-sensor visibility}
\label{subsec:alfred_fov}

The two experiments place different demands on the frame-mounted LiDAR. In ForestYear3D, the robot traverses a forest corridor with the vegetation of interest to either side, and an operator steers, so transverse visibility matters more than longitudinal visibility. In ROWCAST, ALFRED~2.0 drives autonomously. An Ackermann-steered vehicle can only move along its longitudinal axis, with steering changing only the curvature, so self-occlusion to the front or rear hides the region the platform will enter next and matters more than the same occlusion at an oblique bearing.

\begin{figure}[htbp]
    \centering
    \includegraphics[width=0.2\linewidth]{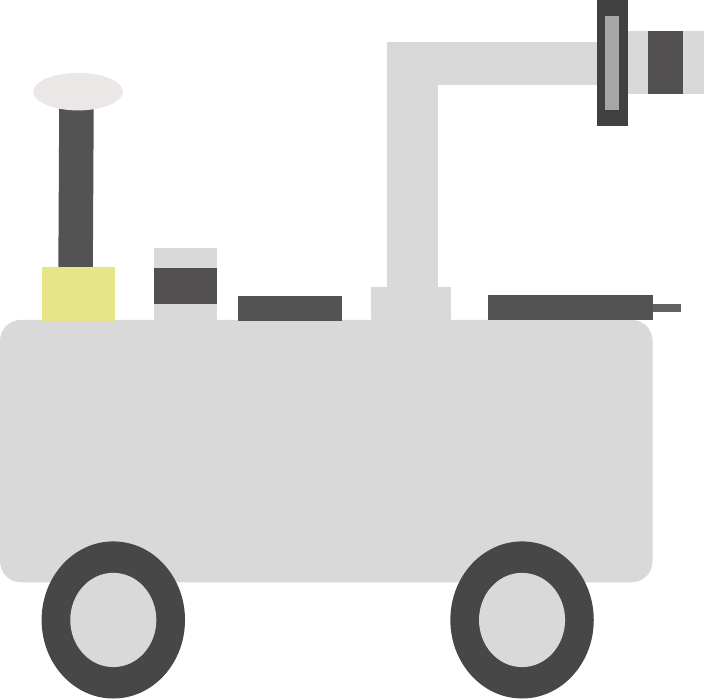}
    \caption{ALFRED in the drive posture, also used as the angle-scan posture in ForestYear3D.}
    \label{fig:drive_pos}
\end{figure}

Visibility was evaluated for three beams, as emitted in the sensor frame: the lowest (\SI{-22.5}{\degree}), the one nearest horizontal (\SI{-0.7258}{\degree}) and the highest (\SI{+22.5}{\degree}). Rays were cast at 1,024 azimuths up to a \SI{5}{\metre} cap and intersected with an analytical ground plane and the posed visual meshes, rather than the simplified collision primitives used for planning. Azimuth \SI{0}{\degree} is the robot's forward direction. Each build was evaluated with the manipulator at zero (see Figure~\ref{fig:alfred_sim_builds}) and in the drive posture $q_{\mathrm{drive}} = [1.5,\ 1.5,\ -1.5,\ 0,\ 0,\ 0]$ (see Figure~\ref{fig:drive_pos}), which served as the angle-scan posture in ForestYear3D and the drive posture in crop-row operation.

\begin{figure}[htbp]
    \centering
    \includegraphics[width=0.7\linewidth]{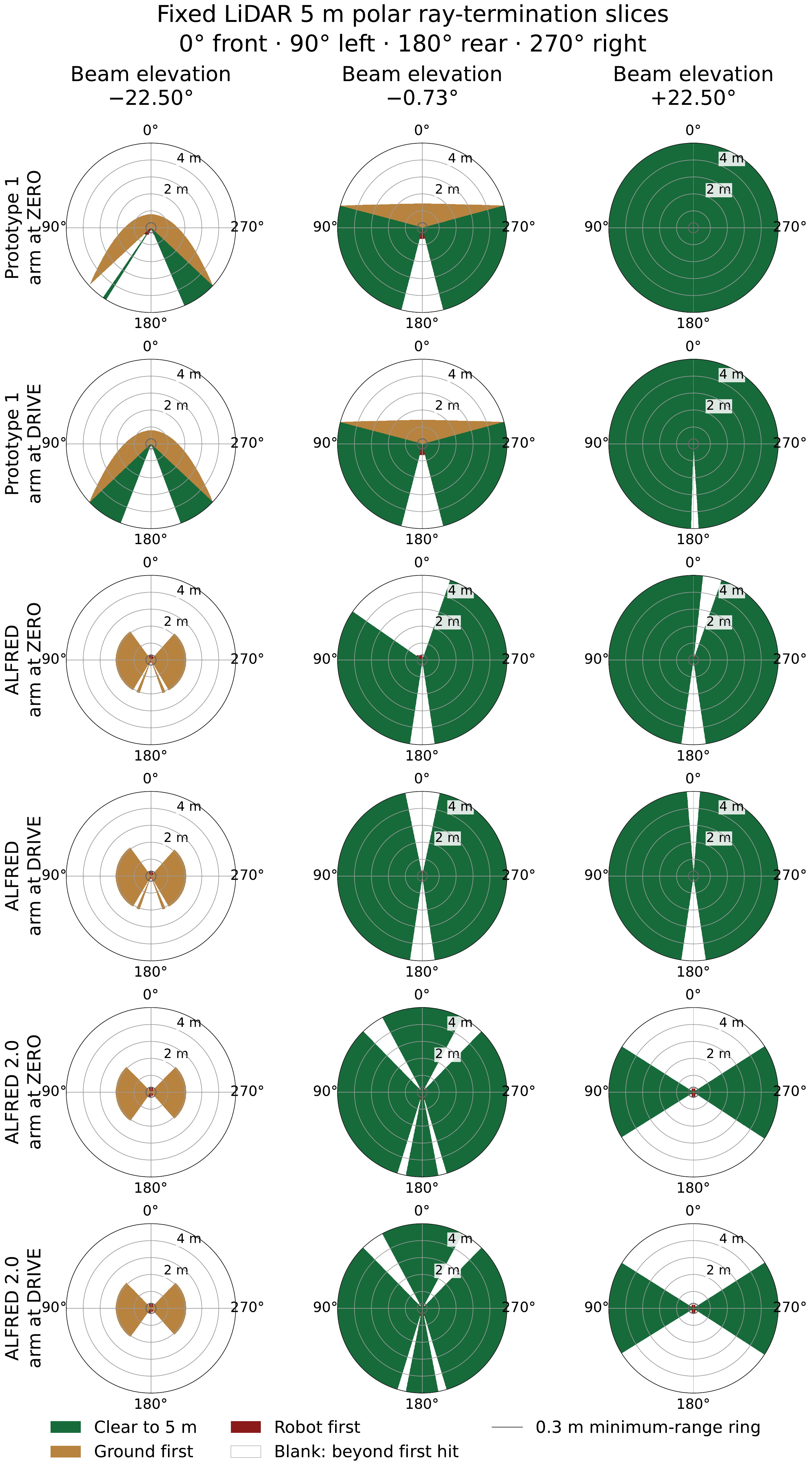}
    \caption{Ray termination for the \SI{-22.5}{\degree}, \SI{-0.7258}{\degree} and \SI{+22.5}{\degree} beams. Green rays reach the \SI{5}{\metre} cap, ochre rays meet the ground first and red rays meet the platform. White marks the region beyond a first intersection, not absent coverage.}
    \label{fig:alfred_fov_polar}
\end{figure}

\begin{table}[htbp]
    \centering
    \caption{Self-occlusion of the frame-mounted LiDAR for each build and arm posture. \emph{Clear aperture} is the clear arc about the forward/rear (longitudinal) and left/right (transverse) directions; \emph{scan share} is the percentage of azimuths reaching the \SI{5}{\metre} cap (clear) or terminating on the platform; \emph{largest arc} is the widest contiguous platform-occluded and non-clear sector. $^{\dagger}$Prototype~1's sensor is pitched. $^{\ddagger}$Beam elevations are as emitted in the sensor frame.}
    \label{tab:alfred_fov_full}
    \footnotesize
    \setlength{\tabcolsep}{2pt}
    \begin{tabular}{@{}>{\raggedright\arraybackslash}p{2.1cm}?r?cc?cc?cc@{}}
        \toprule
        & Beam & \multicolumn{2}{?c}{Clear aperture [deg]} & \multicolumn{2}{?c}{Scan share [\%]} & \multicolumn{2}{?c}{Largest arc [deg]} \\
        \cmidrule(lr){3-4}\cmidrule(lr){5-6}\cmidrule(lr){7-8}
        Configuration & [deg] & Long.\ F/R & Trans.\ L/R & Clear & Plat. & Plat. & Non-clear \\
        \midrule
        \multirow{3}{2.1cm}{\raggedright Prototype 1$^{\dagger}$, zero} & $-$22.50$^{\ddagger}$ & 0.00/0.00 & 0.00/0.00 & 7.32 & 18.65 & 54.84 & 278.79 \\
         & $-$0.73 & 0.00/0.00 & 90.70/90.70 & 50.39 & 7.91 & 28.48 & 150.12 \\
         & +22.50 & 360.00/360.00 & 360.00/360.00 & 100.00 & 0.00 & 0.00 & 0.00 \\
        \addlinespace[2pt]
        \grouprule
        \multirow{3}{2.1cm}{\raggedright Prototype 1$^{\dagger}$, drive} & $-$22.50$^{\ddagger}$ & 0.00/0.00 & 0.00/0.00 & 14.45 & 11.43 & 41.13 & 266.84 \\
         & $-$0.73 & 0.00/0.00 & 90.70/90.70 & 50.39 & 7.91 & 28.48 & 150.12 \\
         & +22.50 & 355.08/0.00 & 355.08/355.08 & 98.63 & 1.37 & 4.92 & 4.92 \\
        \addlinespace[2pt]
        \grouprule
        \multirow{3}{2.1cm}{\raggedright ALFRED, zero} & $-$22.50$^{\ddagger}$ & 0.00/0.00 & 0.00/0.00 & 0.00 & 35.55 & 77.34 & 360.00 \\
         & $-$0.73 & 0.00/0.00 & 116.72/152.58 & 74.80 & 25.20 & 74.18 & 74.18 \\
         & +22.50 & 178.24/0.00 & 178.24/152.58 & 91.89 & 8.11 & 16.52 & 16.52 \\
        \addlinespace[2pt]
        \grouprule
        \multirow{3}{2.1cm}{\raggedright ALFRED, drive} & $-$22.50$^{\ddagger}$ & 0.00/0.00 & 0.00/0.00 & 0.00 & 35.55 & 77.34 & 360.00 \\
         & $-$0.73 & 0.00/0.00 & 160.31/159.96 & 88.96 & 11.04 & 23.20 & 23.20 \\
         & +22.50 & 0.00/0.00 & 167.34/167.34 & 92.97 & 7.03 & 16.52 & 16.52 \\
        \addlinespace[2pt]
        \grouprule
        \multirow{3}{2.1cm}{\raggedright ALFRED 2.0, zero} & $-$22.50$^{\ddagger}$ & 0.00/0.00 & 0.00/0.00 & 0.00 & 45.90 & 88.24 & 360.00 \\
         & $-$0.73 & 56.60/22.50 & 118.83/119.18 & 88.09 & 11.91 & 15.82 & 15.82 \\
         & +22.50 & 0.00/0.00 & 64.69/66.09 & 36.33 & 63.67 & 114.61 & 114.61 \\
        \addlinespace[2pt]
        \grouprule
        \multirow{3}{2.1cm}{\raggedright ALFRED 2.0, drive} & $-$22.50$^{\ddagger}$ & 0.00/0.00 & 0.00/0.00 & 0.00 & 45.90 & 88.24 & 360.00 \\
         & $-$0.73 & 56.60/22.50 & 118.83/119.18 & 88.09 & 11.91 & 15.82 & 15.82 \\
         & +22.50 & 0.00/0.00 & 64.69/66.09 & 36.33 & 63.67 & 114.61 & 114.61 \\
        \bottomrule
    \end{tabular}
\end{table}

The results are shown in Figure~\ref{fig:alfred_fov_polar} and Table~\ref{tab:alfred_fov_full}. On ALFRED, arm posture matters: the clear share of the horizontal beam rises from 74.80\% with the arm at zero to 88.96\% in the drive posture. On ALFRED~2.0 the two postures give identical results, because the arm no longer crosses the sensor's horizontal plane, and removing this dependence on posture is useful in its own right. Prototype~1's sensor is pitched, so its near-horizontal beam meets the ground across the whole forward sector at \SI{1.41}{\metre}. Its results are a mount-specific reference rather than a like-for-like comparison, showing that an inclined mount gives up visibility in the lower beams for clearer visibility in the upper ones.

For ForestYear3D, where transverse visibility matters most, no build's lowest beam reaches the \SI{5}{\metre} cap. In the horizontal beam, ALFRED in the drive posture used during the campaign gives the widest transverse aperture of any configuration, at \SI{160.31}{\degree} to the left and \SI{159.96}{\degree} to the right. ALFRED also far outperforms ALFRED~2.0 in the highest beam (92.97\% clear against 36.33\%), which matters because ForestYear3D aims to capture as much of each tree's height as possible, although Prototype~1's pitched sensor performs best in this beam.

For ROWCAST, the key measure is horizontal-beam visibility along the longitudinal axis. ALFRED~2.0 is the only build whose horizontal beam reaches the cap in these directions, with \SI{56.60}{\degree} of clear aperture to the front and \SI{22.50}{\degree} to the rear. This is a redistribution rather than a uniform gain: ALFRED~2.0's remaining occlusions flank the longitudinal axis rather than crossing it, and its upper layer trades upper-beam coverage for centreline visibility. The analysis is static and assumes flat ground, so it locates the self-occluded regions but not how long an obstacle stays hidden while the robot moves. Answering that would require simulating the sensor's view through realistic turns at a range of speeds and steering angles.

\subsection{Manipulator workspace}
\label{subsec:alfred_workspace}

\begin{figure}[htbp]
    \centering
    \includegraphics[width=0.7\columnwidth]{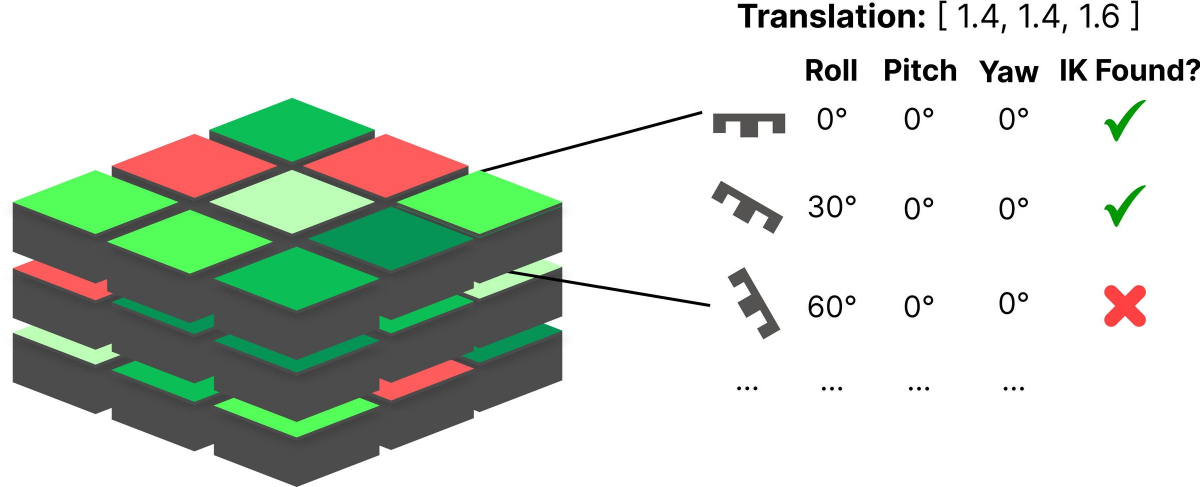}
    \caption{Reachability lattice, with each translation coloured by the number of orientations with a valid inverse kinematics solution.}
    \label{fig:arm_library}
\end{figure}

To test the manipulator's ability to place sensors at poses the frame-mounted sensors cannot reach, each build was evaluated on a common lattice of \num{3110400} end-effector poses: 1,800 translations at \SI{0.2}{\metre} spacing relative to the manipulator base, each with 1,728 orientations sampled at \SI{30}{\degree} intervals in roll, pitch and yaw (Figure~\ref{fig:arm_library}). Because translations are expressed relative to the manipulator base rather than the vehicle, all three builds are tested at identical poses relative to the arm.

Inverse kinematics was attempted for each target from eight fixed seeds. A solution was accepted only if it converged to within \SI{2}{\milli\metre} and \SI{0.5}{\degree} and was collision-free under the build's own collision model. Because the manipulator uses relative encoders, its initial pose can be slightly off, so two safety margins were added: a \SI{0.10}{\metre} clearance between the arm and the rest of the robot, and joint limits inset by \SI{0.23}{\radian}. As the search is finite, a failure means no admissible configuration was found within the budget, not that none exists, so absolute figures are lower bounds. The comparison holds because the lattice, budget and tolerances are identical across builds.

\begin{table}[htbp]
    \centering
    \caption{Usable manipulator workspace by build, from \num{3110400} target poses each. \emph{Pairs} is the number of collision pairs checked; \emph{reachable} counts poses at which inverse kinematics converged within the inset joint limits, of which collision and clearance rejections are mutually exclusive subsets. Percentages are of reachable poses.}
    \label{tab:alfred_reachability_results}
    \small
    \begin{tabular}{@{}lrrrrrr@{}}
        \toprule
        Build & Pairs & Reachable & Collision & Clearance & Usable & Usable [\%] \\
        \midrule
        Prototype 1 & 128 & 25,096 & 12,751 & 3,821 &  8,524 & 34.0 \\
        ALFRED      & 102 & 24,950 &  4,398 & 5,578 & 14,974 & 60.0 \\
        ALFRED 2.0  &  84 & 24,950 &  3,310 & 5,146 & 16,494 & 66.1 \\
        \bottomrule
    \end{tabular}
\end{table}

\begin{figure}[htbp]
    \centering
    \includegraphics[width=0.9\linewidth]{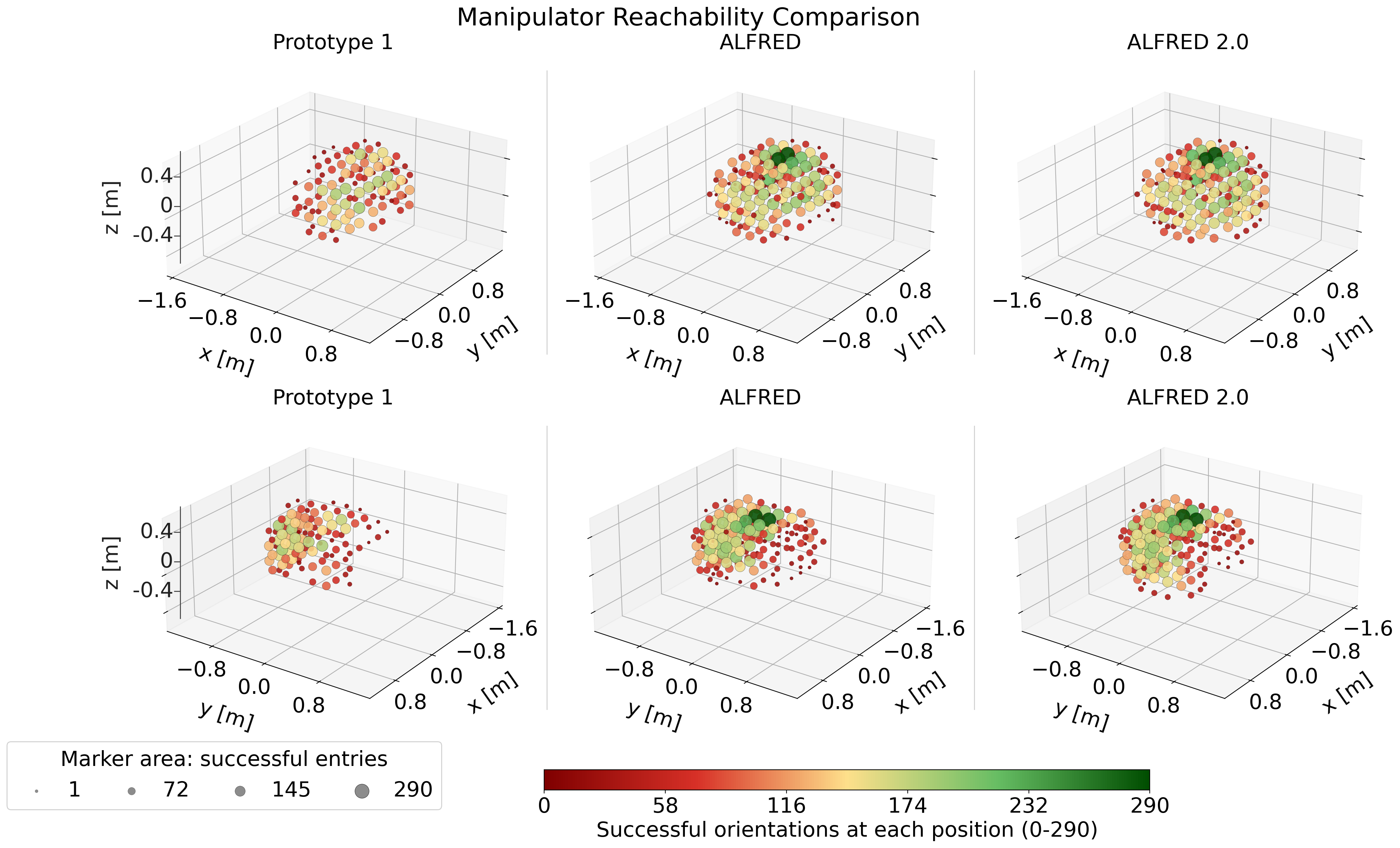}
    \caption{Usable workspace by end-effector position relative to the manipulator base; the lower row repeats the upper at a rotated azimuth. Each marker is a translation cell retaining at least one of the 1,728 orientations, and empty cells are omitted. Marker area and colour scale with the number of retained orientations on a common 0--290 range, so the darkest green is the observed maximum rather than a full set.}
    \label{fig:alfred_reachability_support}
\end{figure}

Most of the lattice lies beyond the manipulator's \SI{740}{\milli\metre} reach, so the informative denominator is the set of poses at which inverse kinematics converged. This set is almost identical across builds (Table~\ref{tab:alfred_reachability_results}); Prototype~1's slightly larger set reflects its different end-effector mount, which places the wrist differently for the same sensor pose. Of these reachable poses, 34.0\% are usable on Prototype~1, 60.0\% on ALFRED and 66.1\% on ALFRED~2.0. The strut frame therefore increased the usable workspace by a factor of about 1.8, and the upper layer added a further six percentage points.

The gain comes from fewer collisions. Collision rejections fall from 50.8\% of reachable poses on Prototype~1 to 17.6\% on ALFRED and 13.3\% on ALFRED~2.0, while the number of collision pairs falls from 128 to 102 to 84, showing that structure was removed from around the arm rather than rearranged. Figure~\ref{fig:alfred_reachability_support} shows the same trend spatially: Prototype~1 retains at least one orientation at 103 of the 1,800 translation cells, against 157 and 161 for the later builds, and its best cell admits 184 orientations against 290. The later builds therefore reach more positions and admit more orientations at each.

Prototype~1's cluttered frame limits its usable workspace to 34\% of reachable poses despite moderate visibility, so it partially satisfies R4. ALFRED and ALFRED~2.0 both perform well in reachability, and each has visibility suited to its experiment, so both satisfy R4.

\subsection{R5: Endurance}
\label{sec:alfred_r5}

Prototype~1 was never tested in the field, but it used the same two independently powered enclosures later deployed on ALFRED, so it receives the same rating: partially satisfied. ALFRED only partially satisfies R5 because, around December 2025, one of its two batteries degraded significantly. Each charge then lasted about \SI{30}{\minute}, or roughly five of the 33 corridor traversals in a complete survey window, so a window that had taken one day to collect took three. Winter made this worse, since trials cannot run in rain or strong wind and collection days were already scarce. A replacement \SI{48}{\volt} \SI{30}{\ampere\hour} lithium-ion battery and a portable generator were obtained, so R5 was restored by new equipment part-way through the campaign rather than by a new build. ALFRED~2.0 used both from the start and satisfies R5.

\subsection{R6: Reproducibility by others}
\label{sec:alfred_reproducibility}

The platform is released in full at \url{https://github.com/CiaranJohnson/ALFRED}. The release includes the source CAD for every bespoke component; the robot descriptions and collision configurations used in Section~\ref{sec:alfred_reachability}; and code for operating the platform in simulation and on hardware. The software is packaged in Docker containers so it runs independently of the host environment, an implementation manual covers assembly and bring-up, and the bill of materials in \ref{sec:alfred_appendix} lists every commercial component with its cost. Prototype~1's robot description is included, but as that build was never deployed, no operating code exists for it, so it partially satisfies R6. ALFRED and ALFRED~2.0 are fully reproducible from the release and satisfy R6.

\section{Discussion, Limitations and Research Opportunities}
\label{sec:alfred_discussion}

The requirements set out in Section~\ref{sec:alfred_requirements} were met in service rather than in demonstration. ALFRED completed a year of monthly forest surveys, 528 traversals across 29 days, without missing a scheduled collection, while the platform was also reconfigured for a concurrent payload study and while ALFRED~2.0 was being built and tested alongside it. Iterating against explicit requirements is what made that possible: each rebuild was directed at a specific shortcoming of its predecessor, and the effect is measurable. The usable share of the arm's reachable workspace rose from 34.0\% to 60.0\% and then 66.1\%, the collision pairs the model must check fell from 128 to 84, and only the final build keeps the frame-mounted LiDAR's view clear along both directions of travel. Because both comparisons were made on the robot descriptions, the consequence of a proposed change can be quantified before anything is manufactured. This is the most transferable result of the work, and with the design files, robot description and containers released, the same comparison can be run on other platforms. The remainder of this section sets out what the deployment record shows about the limits inherited from the commercial hardware, the integration effort the platform required, and the proposing future work.

\subsection{Hardware}
There are two major limitations firstly the chassis is rated to IP22 unless purchased with custom ingress protection, and the manual states explicitly that it is not waterproof and should not be operated in rain or snow; this is why collection was not attempted in wet conditions. Secondly, the arm uses relative encoders and must be placed in the zero position prior to power on. This results in joint offsets, this error can propagate into sensor measurements.

\subsection{Integration is the hidden cost}
\label{subsec:disc_integration}
The substantive engineering effort lay in making the commercial hardware and open-source software work together. Modifications to third-party code were needed, to integrate the various components of the robot. The most consequential was a mismatch between the LiDAR driver's packet structure and that expected from FAST-LIO2. Another significant modification is the base does not expose a joint state interface on the physical hardware: it reports only a rear-wheel linear velocity and a central steering angle, so the positions of the steering and wheel joints are unknown to the rest of the system and the robot model cannot be posed. Without them there is no complete transform tree, and consequently no way to view the LiDAR and camera output in its correct spatial relationship to the robot; this removes the most direct means of checking, live or from a recording, that the sensors are seeing what they are believed to be seeing. A hardware interface was added within the vendor package to reconstruct the joint states from the reported velocity and steering angle using the platform's Ackermann geometry, resolving per-wheel steering angles from the lateral distance to the instantaneous centre of rotation and per-wheel speeds from their respective path radii. This also reconciles the simulator, whose model is Ackermann, with the hardware, which reports according to a bicycle model, so that one visualisation serves both.

\subsection{Future Work}
\label{subsec:future_work}
ALFRED~2.0 was initially intended to support online next-best-view planning in which the robot would observe the current scene, select an informative sensor pose, generate a collision-free whole-body motion and execute that motion before repeating the process. Such an architecture would allow the robot to adapt its sensing trajectory to the vegetation present during each visit rather than relying on viewpoints selected in advance. Several planning and control approaches were implemented towards this objective.

The initial architecture adapted an existing mobile-manipulator next-best-view framework to the ALFRED platform. A live occupancy representation was constructed from the frame-mounted and arm-mounted LiDAR measurements. Candidate sensor poses were sampled around the observed scene and evaluated by casting rays into the occupancy map to estimate how much previously unobserved space would become visible. Candidate poses were then tested for inverse-kinematic feasibility, self-collision and collision with the mapped environment. The intended output was a sensor pose that was both reachable and expected to increase the observed volume.

The motion required to reach the selected pose was represented as a whole-body path containing the planar vehicle pose and the six manipulator joint positions. Early implementations used sampling-based planning with Reeds--Shepp connections for the base and separate inverse kinematics for the manipulator. A subsequent implementation used a local quadratic programme to steer a search tree through the combined base-arm configuration space. The quadratic programme selected a signed base velocity, steering change and six joint velocities to reduce the error to a sampled whole-body state while respecting steering, joint-position and joint-velocity limits. While also being collision-free with itself and the environment. The issue with this approach is the time required to generate the next motion depended strongly on the scene, the sampled candidates and the number of rejected paths and could grow extremely large.

 A kinematic whole-body model-predictive controller (MPC) was developed to track the resulting base-arm trajectories. The vehicle was represented using an Ackermann bicycle model and the manipulator joints were propagated using first-order kinematics. The MPC introduced a segmented path-progress tracker to avoid cutting corners in both the base and the manipulator. However all attempts encountered difficulties at transitions between forward and reverse motion and struggled to generalise well to all scenarios. 

 Online collision checking also depended on constructing a sufficiently consistent representation of the changing scene. ALFRED 2.0 update it's representation of the environment using both the base- and arm-mounted LiDARs. Separate self-filters were required because each sensor observed different parts of the robot. Their filtered measurements were then combined for insertion into an OctoMap used by the planner and collision checker. This pipeline was sensitive to the consistency of the point-cloud timestamp, transform tree and self-filter geometry. Inaccuracies in any of these could result in the robot adding part of itself to the OctoMap. The collision checker could then report a collision between the robot and a structure generated by the robot itself. Increasing the filtered volume reduced this risk but also removed legitimate environmental measurements close to the platform.

Future work, would look to address these issues identified in the online planner and compare this approach to that of ROWCAST for autonomous crop row monitoring. 

\section{Conclusion}
\label{sec:alfred_conclusion}

Repeated in-field sensing demands more of a platform than a single demonstration: it must tolerate the weather, adapt to changing sensing tasks, be repaired between visits and last a full collection session. This paper presented ALFRED, a mobile manipulator assembled from commercially available components and developed through four builds against six requirements set for that purpose. The platform met them in service, completing a year of monthly forest surveys (528 traversals over 29 days, \SI{4.9}{\tera\byte} of data) without missing a scheduled collection, while also being reconfigured for a concurrent payload study and rebuilt for autonomous crop-row operation, with each change of build taking about six hours. Only endurance needed intervention, restored mid-campaign by replacing a degraded battery.

Iterating against explicit requirements produced measurable gains. Across the last three builds, the usable share of the arm's reachable workspace rose from 34.0\% to 60.0\% and then 66.1\%, the number of collision pairs fell from 128 to 84, and only the final build keeps the frame-mounted LiDAR's horizontal view clear both forwards and backwards. Because both analyses run on the robot descriptions, the effect of a design change on reach and visibility can be quantified before anything is manufactured. The experience suggests three priorities for similar platforms: design for repair from the outset, treat the robot description as part of the hardware configuration, and budget for integration. The design files, robot descriptions, drivers and containers are released so that others can rebuild the platform or run the same comparisons on their own designs. Whether placing sensors with an arm improves what repeated surveys measure is the question the platform was built to answer, and the data it has already collected is where that answer will be found.

\section*{Acknowledgements}
The authors acknowledge the use of the Heriot-Watt University high-performance computing facility, DMOG, and its associated support services in completing this work.

This work could not have been achieved without the \href{https://uoe.sharepoint.com/sites/InformaticsMakerspace}{University of Edinburgh Informatics Makerspace} in Appleton Towers, and the Edinburgh Award. The authors would also like to thank Edinburgh Centre for Robotics, the University of Edinburgh, Heriot-Watt University, and Harper Adams University for their support in this work. Thank you to the \href{https://oceansystemslab.github.io/}{Ocean Systems Lab} at Heriot-Watt Univeristy for providing a workshop for construction.

\section*{CRediT authorship contribution statement}
\textbf{Ciarán Miceal Johnson:} Conceptualization, Methodology, Software, Investigation, Formal analysis, Data curation, Visualization, Writing -- original draft, Writing -- review \& editing.
\textbf{Christopher Quail:} Conceptualization, Investigation, Writing -- review \& editing.
\textbf{Garry Ellard: } Conceptualization, Investigation, Writing -- review \& editing.
\textbf{Alistair McConnell:} Resources, Supervision, Writing -- review \& editing.
\textbf{Steve Tonneau:} Resources, Supervision, Writing -- review \& editing.
\textbf{Fernando Auat Cheein:} Resources, Supervision, Writing -- review \& editing.

\section*{Data availability}
The design files, robot descriptions, and containerised software are available at \url{https://github.com/CiaranJohnson/ALFRED}. The ForestYear3D dataset is described at \url{https://ciaranjohnson.github.io/ForestYear3D/}, with code at \url{https://github.com/CiaranJohnson/ForestYear3D}. The ROWCAST experiment is described at \url{https://ciaranjohnson.github.io/ROWCAST/}.

\section*{Declaration of competing interests}
The authors declare that they have no known competing financial interests or personal relationships that could have appeared to influence the work reported in this paper.

\section*{Declaration of generative AI and AI-assisted technologies in the manuscript preparation process.}
During the preparation of this work the authors used Claude Opus 5 in order to improve the grammar, structure and coherence of the manuscript. After using this tool, the authors reviewed and edited the content as needed and take full responsibility for the content of the published article.

\section*{Funding}
CJ: PhD funding is provided by Engineering and Physical Sciences Research Council (EPSRC) Centre for Doctoral Training in Robotics and Autonomous Systems. \\

\appendix
\section{Bill of Materials}
\label{sec:alfred_appendix}

\begin{table*}
    \centering
    \caption{Hardware components of the ALFRED platform. Costs are estimates for the quantity listed, in GBP as purchased between 2023 and 2026, and will vary with supplier and date. The Alienware laptop is the operator's field-edge machine and is not carried on the platform.}
    \resizebox{\textwidth}{!}{%
    \begin{tabular}{||l|l|l|l|r||}
    \hline
    Hardware Type & Company & Device Name & Quantity & Cost (£) \\
    \hline
    \hline
    Mobile Base & Agile X & Hunter 2.0 & 1 & 9,524.44 \\
    \hline
    Robot Arm & Unitree & Z1 & 1 & 11,812.52 \\
    \hline
    Laptop & Dell & Latitude 7330 Rugged Extreme & 1 & 3,992.08 \\ 
    \hline
    Laptop & MSI & Stealth GS77 & 1 & 2,455.00 \\ 
    \hline
    Laptop & Alienware & m16 r1 & 1 & 2,613.44 \\ 
    \hline
    LiDAR & Ouster & OS1-32 &  2 & 10,622.10 \\
    \hline
    RTK GNSS & Swift Nav & Duro & 2 & 4,681.18 \\
    \hline
    RGB-D Camera & Intel Realsense & D457 & 3 & 1,107.06 \\
    \hline
    IMU & InertialSense & RUG-3-IMX-5-RTK & 1 & 1,995.97 \\
    \hline
    Ethernet Hub & NETGEAR & GS305 & 1 & 14.99 \\
    \hline
    Ethernet Hub & NETGEAR & GS308 &  1 & 21.99 \\
    \hline
    Wi-Fi Router & GL.iNet  & GL-AXT1800 Slate AX &  1 & 107.99 \\
    \hline
    USB Hub & RSHTECH & USB Hub Powered & 1 & 19.99 \\
    \hline
    Bluetooth Controller & EasySMX & X15 &  1 & 28.96 \\
    \hline
    Ethernet Cables & RS PRO & 1m Cat6 & 5 & 18.85 \\
    \hline
    USB-C Cables & RØDE & SC27 & 3 & 66.00 \\
    \hline
    Portable Power Station  & Bluetti & AC180 & 1 & 599.00 \\
    \hline
    \hline
    \multicolumn{4}{||l|}{Total} & 49,681.56 \\
    \hline
    \multicolumn{4}{||l|}{Total excluding operator laptop} & 47,068.12 \\
    \hline
    \end{tabular}%
    }
    \label{tab:hardware_table}
\end{table*}

\begin{table}[]
    \centering
    \caption{Main components and estimated costs for the two electronics enclosures, which power the arm, sensors and network switches. Costs are estimates for the quantity listed, in GBP as purchased between 2023 and 2026, and will vary with supplier and date. Fuses, wiring, DC power connectors, threaded connectors and mounting plates are excluded and should be sourced to suit the build.}
    \begin{tabular}{||l|l|l||}
    \hline
    Hardware Type & Amount & Cost (£) \\
    \hline
    \hline
    Waterproof Box & 2 & 54.62 \\
    \hline
    Cable Glands & 6 & 9.99 \\
    \hline
    48V 30mAh Battery & 2 & 158.06 \\
    \hline
    48V to 24V 25A step-down converter & 1 & 41.75 \\
    \hline
    48V to 24V 5A step-down converter & 2 & 84.46 \\
    \hline
    48V to 5V 2A step-down converter & 1 & 9.34 \\
    \hline
    24V to 12V 3A step-down converter & 1 & 11.71 \\
    \hline
    24V to 5V 1A step-down converter & 1 & 8.90 \\
    \hline
    24V to 5V 2A step-down converter & 1 & 8.89 \\
    \hline
    Emergency stop button & 1 & 37.49 \\
    \hline
    Switches & 2 & 2.69 \\
    \hline
    \hline
    Total & & 427.90 \\
    \hline
    \end{tabular}
    \label{tab:electronics_table}
\end{table}

\begin{table}[]
    \centering
    \caption{Main components and estimated costs for the aluminium strut frame and its mounting hardware, covering both the ALFRED frame and the second layer added for ALFRED~2.0. Costs are estimates for the quantity listed, in GBP as purchased between 2023 and 2026, and will vary with supplier and date. Filament (PLA, PETG, TPU), common fasteners and access to a 3D printer and laser cutter are assumed and excluded.}
    \begin{tabular}{||l|l|l||}
    \hline
    Hardware Type & Amount & Cost \\
    \hline
    \hline
    40x40mm 8mm Groove & 10,000 mm & 330.25 \\
    RS Pro Profile Strut & & \\

    \hline
    40x40x40mm 8mm Groove & 72 & 662.40 \\
    RS Pro Angle Brackets & & \\

    \hline
    3mm Acrylic & 270x996 mm  & 14.06 \\
    \hline
    6mm Acrylic & 195x1836 mm & 51.16 \\
                & 150x290 mm  &       \\
    \hline
    12mm Acrylic & 160x465mm & 8.90 \\
    \hline
    Threaded Inserts & 10 & 14.58 \\
    \hline
    Bolt and latch & 2 & 22.76 \\
    \hline
    Door Hinge & 4 & 61.16 \\
    \hline
    \hline
    Total & & 1,161.86 \\
    \hline
    \end{tabular}
    \label{tab:strut_frame_table}
\end{table}

 \bibliographystyle{elsarticle-num-names} 
 \bibliography{cas-refs}

\end{document}